\documentclass[letterpaper, 10 ptjournal, twoside]{ieeetran}

\IEEEoverridecommandlockouts

\pdfoutput=1
\usepackage{cite}
\usepackage{amsmath,amssymb,amsfonts}
\usepackage{algorithmic}
\usepackage{graphicx}
\usepackage{textcomp}
\usepackage{xcolor}
\usepackage{siunitx}
\usepackage{booktabs}
\usepackage{tabularx}
\usepackage{multirow}
\usepackage{array}
\usepackage{makecell}
\usepackage{amsmath}
\usepackage{lineno} 
\usepackage{caption}
\usepackage[font=footnotesize]{caption}
\usepackage{soul}
\usepackage{algorithm}
\usepackage{algorithmic}

\usepackage{makecell}

\soulregister{\cite}7 
\soulregister{\citep}7 
\soulregister{\citet}7 
\soulregister{\ref}7 
\soulregister{\pageref}7 
\soulregister{\eqref}7 
\soulregister{\ang}7
\usepackage[caption=false,font=footnotesize,labelfont=rm,textfont=rm]{subfig}

\usepackage{tikz,xcolor}
\usepackage[hidelinks]{hyperref}

\definecolor{highlightcolor}{RGB}{255,0,0}

\definecolor{lime}{HTML}{A6CE39}
\DeclareRobustCommand{\orcidicon}{%
    \begin{tikzpicture}
    \draw[lime, fill=lime] (0,0) 
    circle [radius=0.16] 
    node[white] {{\fontfamily{qag}\selectfont \tiny ID}};    \draw[white, fill=white] (-0.0625,0.095) 
    circle [radius=0.007];    \end{tikzpicture}
    \hspace{-2mm}}
\foreach \x in {A, ..., Z}{%
    \expandafter\xdef\csname orcid\x\endcsname{\noexpand\href{https://orcid.org/\csname orcidauthor\x\endcsname}{\noexpand\orcidicon}}
    }

\hypersetup{
    colorlinks=False,
    linkcolor=blue, 
    urlcolor=black 
    citecolor=blue, 
    filecolor=black, 
    menucolor=black 
}

\def\BibTeX{{\rm B\kern-.05em{\sc i\kern-.025em b}\kern-.08em
    T\kern-.1667em\lower.7ex\hbox{E}\kern-.125emX}}

\makeatletter
\def\sf@counterlist{}
\makeatother

\usepackage{xpatch}

\makeatletter
\xpatchcmd{\IEEEbiography}{plus 1fil}{}{}{}
\xpatchcmd{\endIEEEbiography}{plus 1fil}{}{}{}
\xpatchcmd{\IEEEbiographynophoto}{plus 1fil}{}{}{}
\xpatchcmd{\endIEEEbiographynophoto}{plus 1fil}{}{}{}

\def\@IEEEBIOskipN{4\baselineskip}
\makeatother

\begin{document}

\title{Bi-Level Reinforcement Learning Control for an Underactuated Blimp via Center-of-Mass Reconfiguration\\
\author{Xiaorui Wang\textsuperscript{$\dagger$}, Hongwu Wang\textsuperscript{$\dagger$}, Yue Fan, Hao Cheng and Feitian Zhang\textsuperscript{*}}

\thanks{\textsuperscript{$\dagger$} These authors contributed equally to this work.

The authors are with the Robotics and Control Laboratory, School of Advanced Manufacturing and Robotics, and the State Key Laboratory of Turbulence and Complex Systems,  Peking University, Beijing, 100871, China (email: \href{mailto:jnswxr@stu.pku.edu.cn}{jnswxr@stu.pku.edu.cn}; \href{mailto:whw@stu.pku.edu.cn}{whw@stu.pku.edu.cn}; \href{mailto:2301213157@stu.pku.edu.cn}{2301213157@stu.pku.edu.cn}; \href{mailto:h-cheng@stu.pku.edu.cn}{h-cheng@stu.pku.edu.cn}; \href{mailto:feitian@pku.edu.cn}{feitian@pku.edu.cn}).
}

}

\maketitle
\pagestyle{empty}  
\thispagestyle{empty} 

\begin{abstract}
This paper investigates goal-directed tracking control of  underactuated blimps with center-of-mass (CoM) reconfiguration. Unlike conventional overactuated blimp designs that rely on redundant actuation for simplified control, this paper focuses on a compact architecture consisting of two thrusters and a movable internal slider, aiming to improve energy efficiency and payload capacity. This hardware-efficient configuration introduces significant underactuation and strong nonlinear coupling between CoM dynamics and vehicle motion. To address these challenges, this paper proposes a bi-level reinforcement learning framework that explicitly decouples task-level CoM planning from continuous thrust control. The outer policy determines a target-dependent CoM configuration prior to flight, while the inner policy generates thrust commands to track straight-line references. To ensure stable learning, this paper introduces a two-stage learning strategy, supported by a convergence analysis of the resulting bi-level process. 
Extensive simulations and real-world experiments on a 27-goal evaluation set demonstrate that the proposed method consistently outperforms fixed-CoM baselines and PID-based controllers, achieving higher tracking accuracy, enhanced robustness, and reliable sim-to-real transfer.
\end{abstract}

\begin{IEEEkeywords}
Blimp Control, Underactuated Systems, Hierarchical Reinforcement Learning, Center-of-Mass Reconfiguration, Sim-to-Real Transfer
\end{IEEEkeywords}

\section{Introduction}
\IEEEPARstart{L}{ighter}-than-air (LTA) aerial vehicles have attracted increasing attention due to their low power consumption, extended endurance, and inherently safe operation in cluttered or human-centered environments \cite{TIE-blimp-Control-2025, TRO-2022-review, TASE-2025-review}. These characteristics make robotic blimps well-suited for persistent sensing, inspection, search and rescue, and indoor service tasks where quiet flight and long-duration flight are critical \cite{RAL-env-sense-2024, IROS-2017-interact, Blimp-review-2024}. Consequently, LTA platforms have emerged as a key direction for energy-efficient aerial autonomy \cite{RAL-2018-sense, science-2022-interact}.

To enhance maneuverability, existing blimps often employ multiple distributed actuators, typically comprising four or more propulsion units \cite{TIE-blimp-Control-2025, TASE-2025-review, SBlimp-IROS-2023, Tmech-2021-swing-reducing}. While such configurations improve control authority, they inevitably introduce additional mass, increased mechanical complexity, and higher power consumption \cite{RAL-env-sense-2024, TRO-RGBlimp-Q, ICSL-2024-feedback-linear}. For LTA systems, these trade-offs are particularly restrictive, as actuation overhead directly constrains endurance and payload capability. 
This has motivated the development of minimally actuated and structurally reconfigurable designs that exploit morphology rather than actuator redundancy for control. 

A representative direction is the class of underactuated  blimps with CoM reconfiguration, which employ limited thrust actuation combined with movable mass for attitude and trajectory modulation.
While this design improves hardware efficiency, it fundamentally alters the system dynamics by introducing strong nonlinear coupling between translational motion, attitude evolution, and buoyancy-induced effects. Conventional control methods \cite{Ocean-2024-PID, Wan-2018-PID, lopez-2009-PID}, including PID, feedback linearization, and backstepping, rely heavily on accurate system modeling and extensive tuning, and exhibit degraded performance under severe underactuation \cite{TIE-blimp-Control-2025, ICSL-2024-feedback-linear, NSTSMC-2023}. Moreover, CoM modulation is often treated as an external disturbance rather than as an explicit control degree of freedom, limiting its potential for task-oriented optimization \cite{Tmech-2021-swing-reducing, Tmech-2022-fixed-wing}.

\begin{figure*}[tbp]
\centerline{\includegraphics[width=1.0\linewidth]{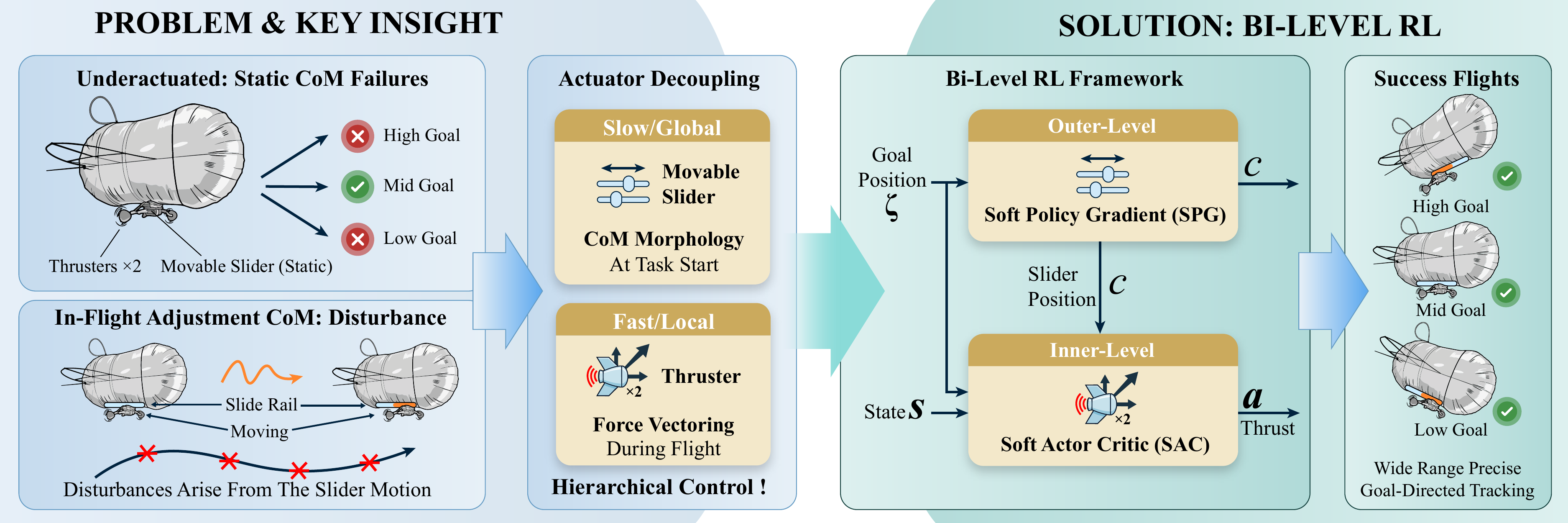}}
\caption{Problem setting, key insight, and proposed bi-level RL solution for goal-directed tracking control of RGBlimp. The platform is severely underactuated, with only two thrusters and a movable slider for CoM reconfiguration. A fixed slider configuration is insufficient for different goals, whereas in-flight slider motion introduces additional disturbance due to internal relative dynamics. The key idea is to decouple the slow task-level CoM adjustment from fast thrust control. Accordingly, a bi-level reinforcement learning framework is proposed. The outer level uses Soft Policy Gradient to determine the slider position prior to flight, and the inner level uses Soft Actor-Critic to generate continuous thruster commands during flight. This hierarchical design enables stable and precise goal-directed tracking over a wide range of target positions.}
\label{fig-intro}
\end{figure*}

Reinforcement learning (RL) offers a data-driven alternative capable of handling nonlinear dynamics and model uncertainty \cite{TIE-2026-RL, TIE-2025-RL, TIE-RL-2025-helicopter}. Prior work has demonstrated promising results in blimp stabilization and waypoint tracking \cite{Blimp-review-2024, Hinf-RL-2023}. However, directly applying a  monolithic RL policy is suboptimal for underactuated blimp systems with reconfigurable mass distribution. In particular, CoM adjustment and thrust generation operate on distinct time scales and induce qualitatively different dynamical responses. In-flight mass redistribution introduces transient disturbances that complicate stabilization and control, as illustrated in Fig.~\ref{fig-intro}, thereby leading to increased training  difficulty and degraded convergence behavior when learned jointly.

This observation motivates a hierarchical formulation of the control problem. Hierarchical and bi-level RL methods decompose decision making across distinct functional or temporal levels \cite{TIE-bi-GTHSL-2025, TNNLS-2022-bi}. While such approaches in robotics have predominantly focused on navigation and planning \cite{TIE-2021-bi, RAL-2021-bi}, the underlying principle is well aligned with underactuated blimps with CoM adjustment, where configuration planning and continuous control are naturally decoupled.

In this work, we propose a bi-level RL framework for  underactuated blimps with CoM reconfiguration. Using a representative RGBlimp platform \cite{RGBlimp-RAL}, we address goal-directed tracking as a coupled configuration-control problem. The outer policy optimizes the CoM configuration by selecting the slider position conditioned on the target, thereby shaping the system dynamics prior to flight. The inner policy generates continuous thrust commands to track a straight-line reference trajectory. The inner controller is learned using Soft Actor-Critic (SAC), while the outer policy is optimized via Soft Policy Gradient (SPG). This decomposition explicitly separates configuration planning and continuous control, thereby mitigating in-flight  disturbances induced by mass reconfiguration and improving training efficiency.

To improve optimization stability, we introduce a two-stage training strategy. The first stage pretrains the inner controller under randomized CoM configurations to establish a robust control basis. The second stage then jointly optimizes the outer and inner policies to coordinate CoM selection and thrust control. We further  provide a convergence analysis for the resulting bi-level learning dynamics. For sim-to-real transfer \cite{TIE-RL-2023-shark, TIE-RL-2026-leg, IROS-2024-sim2real}, 
extensive experiments on a real RGBlimp platform across 27 goal locations demonstrate that the proposed method consistently outperforms fixed-configuration and PID-based baselines, achieving improved tracking accuracy and robustness under varying reconfiguration.

The main contributions of this paper are threefold. 
First, we propose a bi-level RL framework for underactuated blimps with CoM reconfiguration, explicitly decoupling task-dependent configuration optimization from continuous thrust control. Second, we develop a two-stage training strategy and provide convergence analysis for the resulting bi-level learning framework, improving optimization stability and tractability. 
Third, we validate the proposed method through comprehensive simulation and real-world experiments, demonstrating successful sim-to-real transfer and consistent performance improvements over classical and fixed-configuration baselines.

\section{Task Description}

\subsection{Platform Overview}
\begin{figure}[tbp]
\centerline{\includegraphics[width=1.0\linewidth]{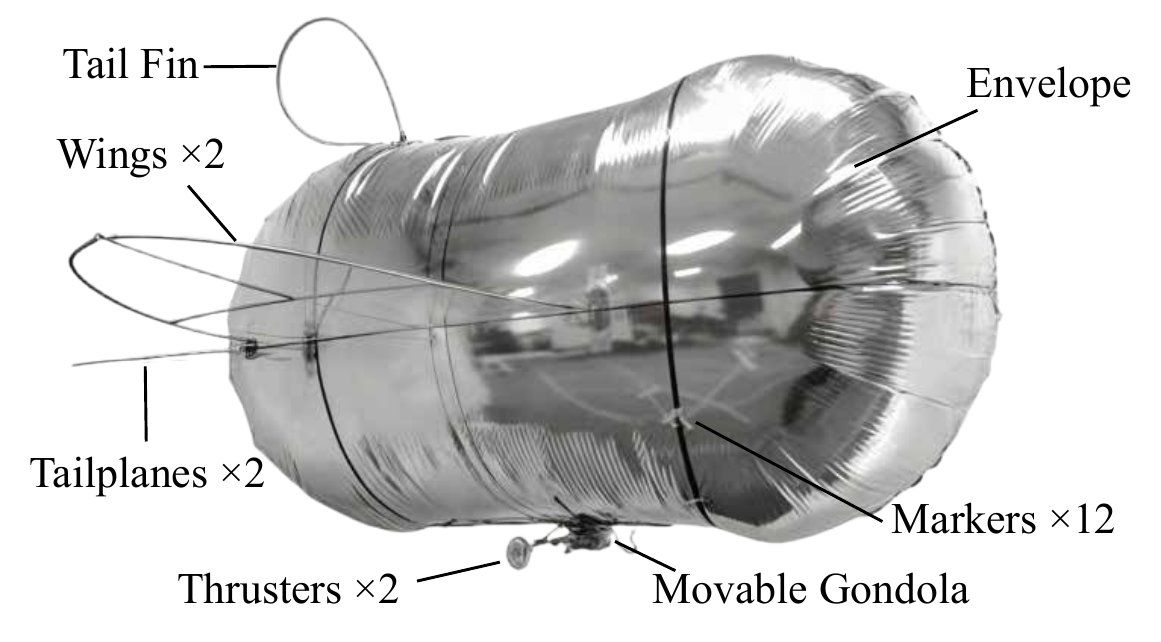}}
\caption{RGBlimp prototype design includes an envelope, a pair of main wings, a tail fin, two tailplanes, and a movable gondola that consists of a pair of propellers, a controller unit, and a battery.}
\label{fig:RGBlimp}
\end{figure}

The RGBlimp is used as a representative underactuated LTA platform in this study. It is equipped with a helium envelope, passive lifting surfaces, two main thrusters, and a longitudinally movable gondola, as illustrated in Fig.~\ref{fig:RGBlimp}. The envelope dimensions are approximately $1.0\times 1.1\times 0.5\,\mathrm{m}$, and the underactuated design is intended to reduce weight, power consumption, and mechanical complexity. The two thrusters provide forward propulsion and differential yaw control, while the movable mass changes the CoM along the body longitudinal axis within a range of $\pm 5\,\mathrm{cm}$. This CoM reconfiguration modifies the equilibrium pitch behavior of the vehicle and provides an additional configuration degree of freedom.

The platform is also equipped with passive aerodynamic surfaces, including main wings and tail structures, to improve low-speed flight stability. For state measurement and experimental evaluation, twelve active LED markers are mounted on the envelope and tracked using an external motion-capture system. Overall, RGBlimp adopts a highly compact and underactuated architecture, enabling lightweight and energy-efficient operation but also  significantly increasing control complexity.

\subsection{Dynamic Modeling}
We adopt a control-oriented 6-DoF model of RGBlimp. Let $\boldsymbol{p}=[x,y,z]^\top$ denote the position in the inertial frame and $\boldsymbol{e}=[\phi,\theta,\psi]^\top$ denote the Euler angles. Let $\boldsymbol{v}_b=[u,v,w]^\top$ and $\boldsymbol{\omega}_b=[p,q,r]^\top$ denote the body-frame linear and angular velocities, respectively. The kinematics are
\begin{equation}
\dot{\boldsymbol{p}}=\boldsymbol{R}(\boldsymbol{e})\,\boldsymbol{v}_b,
\qquad
\dot{\boldsymbol{e}}=\boldsymbol{J}(\boldsymbol{e})\,\boldsymbol{\omega}_b,
\label{eq:rgblimp_kinematics_compact}
\end{equation}
where $\boldsymbol{R}(\boldsymbol{e})$ is the body-to-inertial rotation matrix and $\boldsymbol{J}(\boldsymbol{e})$ is the standard Euler-angle transformation matrix.

Define the generalized velocity and control input as
\begin{equation}
\boldsymbol{\nu}=
\begin{bmatrix}
\boldsymbol{v}_b\\
\boldsymbol{\omega}_b
\end{bmatrix}
\in\mathbb{R}^{6},
\qquad
\boldsymbol{F}=
\begin{bmatrix}
f_l\\
f_r
\end{bmatrix}
\in\mathbb{R}^{2},
\label{eq:rgblimp_nu_input}
\end{equation}
where $f_l$ and $f_r$ represent the left and right thruster forces. Let $c$ denote the slider position. Since the slider is reconfigured before each episode and remains fixed during flight, $c$ is modeled as an episode-wise constant configuration parameter rather than a dynamic state. Under this setting, the blimp dynamics are expressed in a compact state-space form\cite{RGBlimp-RAL}
\begin{equation}
\boldsymbol{M}(c)\dot{\boldsymbol{\nu}}
+\boldsymbol{C}(\boldsymbol{\nu},c)\boldsymbol{\nu}
+\boldsymbol{g}(\boldsymbol{e},c)
+\boldsymbol{d}_a(\boldsymbol{\nu})
=
\boldsymbol{B}(c)\boldsymbol{F}.
\label{eq:rgblimp_dynamics_compact}
\end{equation}
Here, $\boldsymbol{M}(c)$ is the inertia matrix including the effect of the movable mass, $\boldsymbol{C}(\boldsymbol{\nu},c)$ collects the Coriolis and centripetal terms, $\boldsymbol{g}(\boldsymbol{e},c)$ denotes the gravity-buoyancy restoring term induced by the CoM--CoB offset, $\boldsymbol{d}_a(\boldsymbol{\nu})$ represents the aerodynamic forces and moments, and $\boldsymbol{B}(c)$ is the control allocation matrix. The aerodynamic terms depend on the flight state through the angle of attack, the sideslip angle, and the body rates, while the dependence on $c$ captures the effect of CoM configuration on equilibrium and actuation. This model provides the dynamic foundation for the bi-level control formulation presented in the next section. All model parameters were identified from extensive flight experiments under diverse thruster inputs and slider configurations.

\subsection{Goal-Directed Tracking Task}
We consider a goal-directed tracking task in which the blimp starts from the origin and is required to reach a target position $\boldsymbol{\zeta}\in\mathbb{R}^3$ while tracking the straight-line reference from the start to the goal. The target is sampled from the forward workspace, and the reference line is defined as
\begin{equation}
\mathcal{L}(\boldsymbol{\zeta})=
\left\{
\lambda \boldsymbol{\zeta}\mid \lambda\in[0,1]
\right\}.
\label{eq:reference_line}
\end{equation}
The objective is therefore not only to reach the goal but also to minimize deviation from this straight-line path. The corresponding cross-track error is
\begin{equation}
e_{\mathrm{trk}}(\boldsymbol{p}_t,\boldsymbol{\zeta})
=
\min_{\lambda\in[0,1]}
\left\|
\boldsymbol{p}_t-\lambda\boldsymbol{\zeta}
\right\|_2.
\label{eq:tracking_line_error}
\end{equation}


\section{Bi-Level Reinforcement Learning Framework}

\subsection{Hierarchical Optimization Architecture}
Building on the platform model and goal-directed tracking task, we formulate RGBlimp control as a bi-level hierarchical decision-making problem. 
At time step $t$, let $\boldsymbol{s}_t$ denote the system state, $\boldsymbol{a}_t$ the inner-level control action (thruster commands), and $c$ the slider position that reconfigures CoM. 
The task is specified by a target position $\boldsymbol{\zeta}\in\mathbb{R}^3$. 
The inner-level thrust controller is represented by the policy $\pi_{\boldsymbol{\phi}_a}(\boldsymbol{a}_t\mid \boldsymbol{s}_t)$, parameterized by $\boldsymbol{\phi}_a$. 
The outer-level CoM reconfiguration policy is $\pi_{\boldsymbol{\phi}_c}(c\mid \boldsymbol{\zeta})$, parameterized by $\boldsymbol{\phi}_c$, which selects a fixed slider position for the episode.

The augmented state explicitly includes the task specification and CoM configuration, and the inner action is the continuous thrust command
\begin{equation}
\boldsymbol{s}_t \triangleq 
\big[
\boldsymbol{p}_t^{\top},\;
\boldsymbol{e}_t^{\top},\;
\boldsymbol{v}_{b,t}^{\top},\;
\boldsymbol{\omega}_{b,t}^{\top},\;
\boldsymbol{\zeta}^{\top},\;
c
\big]^{\top}
\in \mathbb{R}^{16},
\end{equation}
\begin{equation}
\boldsymbol{a}_t \triangleq 
\big[
f_{l,t},\;
f_{r,t}
\big]^{\top}
\in \mathbb{R}^{2}.
\end{equation}

The hierarchical architecture induces a joint optimization over $\pi_{\boldsymbol{\phi}_c}$ and $\pi_{\boldsymbol{\phi}_a}$. 
For each episode, $\boldsymbol{\zeta}$ is sampled from a task distribution $\mathcal{D}$, the outer policy selects $c\sim \pi_{\boldsymbol{\phi}_c}(c\mid \boldsymbol{\zeta})$, and the inner policy generates a trajectory $\boldsymbol{\tau}$ under the resulting closed-loop dynamics. 
The objective is to maximize the expected discounted return
\begin{equation}
\max_{\boldsymbol{\phi}_c,\, \boldsymbol{\phi}_a}\;
\mathbb{E}_{\boldsymbol{\zeta} \sim \mathcal{D}}
\left[
  \mathbb{E}_{c \sim \pi_{\boldsymbol{\phi}_c}(c \mid \boldsymbol{\zeta})}
  \left[
    \mathbb{E}_{\boldsymbol{\tau} \sim \pi_{\boldsymbol{\phi}_a}}
    \left[
      \sum_{t=0}^{T} \gamma^t\, r(\boldsymbol{s}_t, \boldsymbol{a}_t)
    \right]
  \right]
\right],
\label{eq:joint_objective_no_entropy}
\end{equation}
where $\boldsymbol{\tau}=(\boldsymbol{s}_0,\boldsymbol{a}_0,\boldsymbol{s}_1,\boldsymbol{a}_1,\dots,\boldsymbol{s}_T)$ denotes the trajectory generated by $\pi_{\boldsymbol{\phi}_a}$ with $c$ fixed, $T$ is the episode horizon, $r(\boldsymbol{s}_t,\boldsymbol{a}_t)$ is the per-step reward, and $\gamma\in(0,1)$ is the discount factor.

To optimize the joint objective in~\eqref{eq:joint_objective_no_entropy}, we decompose it into a bi-level optimization with distinct time scales. 
The inner policy $\pi_{\boldsymbol{\phi_a}}(\boldsymbol{a}_t\mid \boldsymbol{s}_t)$ handles step-wise thrust control and is trained using Soft Actor-Critic (SAC), while the outer policy $\pi_{\boldsymbol{\phi}_c}(c\mid \boldsymbol{\zeta})$ selects an episode-wise slider configuration $c$ and is optimized via Soft Policy Gradient (SPG).

\paragraph{Inner Level (SAC)}
We optimize the inner-level thrust controller by maximizing the maximum-entropy return
\begin{equation}
J_{\mathcal{I}}(\boldsymbol{\phi_a}) = 
\mathbb{E}_{\boldsymbol{\tau} \sim \pi_{\boldsymbol{\phi_a}}} \Bigg[
  \sum_{t=0}^{T} \gamma^t \Big(
    r(\boldsymbol{s}_t, \boldsymbol{a}_t) + \alpha \mathcal{H}\big(\pi_{\boldsymbol{\phi_a}}(\cdot \mid \boldsymbol{s}_t)\big)
  \Big)
\Bigg],
\label{eq:inner_obj}
\end{equation}
where $\boldsymbol{\tau} = (\boldsymbol{s}_0, \boldsymbol{a}_0, \boldsymbol{s}_1, \boldsymbol{a}_1, \dots, \boldsymbol{s}_T)$ denotes the induced trajectory, and $\alpha$ is the entropy temperature. SAC is well suited for the inner loop, which involves continuous thruster commands under strongly coupled and underactuated dynamics. The entropy term $\mathcal{H}(\cdot)$ encourages exploration and improves training stability.

Two critic networks $Q_{\boldsymbol{\theta}_i}$, $i\in\{1,2\}$, are learned using transitions sampled from the replay buffer $\mathcal{D}_{R}$
\begin{equation*}
y_t \triangleq r_t + \gamma\,\mathbb{E}_{\boldsymbol{a}_{t+1}\sim \pi_{\boldsymbol{\phi_a}}(\cdot\mid \boldsymbol{s}_{t+1})}\Big[
\min_{j=1,2} Q_{\boldsymbol{\bar{\theta}}_j}(\boldsymbol{s}_{t+1},\boldsymbol{a}_{t+1})
\end{equation*}
\begin{equation}
-\alpha \log \pi_{\boldsymbol{\phi_a}}(\boldsymbol{a}_{t+1}\mid \boldsymbol{s}_{t+1})
\Big],
\label{eq:sac_target}
\end{equation}
\begin{equation}
\mathcal{L}_Q(\boldsymbol{\theta}_i)=
\mathbb{E}_{(\boldsymbol{s}_t,\boldsymbol{a}_t,r_t,\boldsymbol{s}_{t+1})\sim \mathcal{D}_R}
\Big[ \big(Q_{\boldsymbol{\theta}_i}(\boldsymbol{s}_t,\boldsymbol{a}_t)-y_t\big)^2 \Big].
\label{eq:sac_q_loss}
\end{equation}
Target networks are updated using Polyak averaging
\begin{equation}
\boldsymbol{\bar{\theta}} \leftarrow \rho \boldsymbol{\theta} + (1 - \rho)\boldsymbol{\bar{\theta}}, \quad \rho \in (0, 1).
\label{eq:sac_target_update}
\end{equation}
The twin-critic architecture mitigates overestimation bias, and the soft target update improves training smoothness. The actor and temperature are updated by minimizing
\begin{equation*}
\mathcal{L}_{\pi}\left( \boldsymbol{\phi_a} \right) = \mathbb{E}_{\boldsymbol{s}_t \sim \mathcal{D}_{R}}\left[ \mathbb{E}_{\boldsymbol{a}_t \sim \pi_{\boldsymbol{\phi_a}} \left( \cdot \middle| \boldsymbol{s}_t \right)} \left[ \alpha \log \pi_{\boldsymbol{\phi_a}} \left( \boldsymbol{a}_t \middle| \boldsymbol{s}_t \right) 
\right. \right.
\end{equation*}
\begin{equation}
\left. \left.
- \underset{i=1,2}{\min} Q_{\boldsymbol{\theta}_i} \left( \boldsymbol{s}_t, \boldsymbol{a}_t \right) \right] \right],
\label{eq:sac_actor_loss}
\end{equation}
\begin{equation}
\mathcal{L}(\alpha)=
\mathbb{E}_{\boldsymbol{s}_t\sim \mathcal{D}_R,\ \boldsymbol{a}_t\sim \pi_{\boldsymbol{\phi_a}}(\cdot\mid \boldsymbol{s}_t)}
\left[
-\alpha\left(\log \pi_{\boldsymbol{\phi_a}}(\boldsymbol{a}_t\mid \boldsymbol{s}_t)+\hat{H}\right)
\right].
\label{eq:sac_alpha_loss}
\end{equation}
where $\hat{H}$ is the target entropy.

\paragraph{Outer Level (SPG)}
The outer policy $\pi_{\boldsymbol{\phi}_c}(c\mid \boldsymbol{\zeta})$ selects an episode-wise slider configuration $c$ that maximizes the task return achieved by the inner-level controller
\begin{equation}
J_{\mathcal{O}}(\boldsymbol{\phi}_c) = 
\mathbb{E}_{\boldsymbol{\zeta} \sim \mathcal{D}} \left[
  \mathbb{E}_{c \sim \pi_{\boldsymbol{\phi}_c}(\cdot \mid \boldsymbol{\zeta})} \left[
    R(c; \boldsymbol{\zeta}) + \beta \mathcal{H}\!\left(\pi_{\boldsymbol{\phi}_c}(\cdot \mid \boldsymbol{\zeta})\right)
  \right]
\right],
\label{eq:outer_obj}
\end{equation}
where
\begin{equation}
R(c; \boldsymbol{\zeta}) =
\mathbb{E}_{\boldsymbol{\tau} \sim \pi_{\boldsymbol{\phi_a}}}
\left[ \sum_{t=0}^T \gamma^t r(\boldsymbol{s}_t, \boldsymbol{a}_t) \right]
\label{eq:outer_return}
\end{equation}
is the cumulative reward obtained under configuration $c$ using the inner controller $\pi_{\boldsymbol{\phi}_a}$.
SPG is suitable for the outer level because the decision is low-dimensional and executed once per episode. The outer-level entropy term encourages exploration across target-dependent slider configurations. We deliberately exclude the inner-level entropy in $R(c;\boldsymbol{\zeta})$, so that the outer policy evaluates the \emph{actual task reward} obtained under configuration $c$. 
We optimize $\pi_{\boldsymbol{\phi}_c}$ via SPG using stochastic gradient ascent on minibatches, and tune the entropy coefficient $\beta$ by
\begin{equation}
\mathcal{L}(\beta)=
\mathbb{E}_{\boldsymbol{\zeta}\sim \mathcal{D},\ c\sim \pi_{\boldsymbol{\phi}_c}(\cdot\mid \boldsymbol{\zeta})}
\left[
-\beta\left(\log \pi_{\boldsymbol{\phi}_c}(c\mid \boldsymbol{\zeta})+\hat{H}\right)
\right].
\label{eq:spg_beta_loss}
\end{equation}

\subsection{Two-Stage Training}
The bi-level formulation couples the outer CoM reconfiguration policy $\pi_{\boldsymbol{\phi}_c}(c\mid \boldsymbol{\zeta})$ with the inner thrust controller $\pi_{\boldsymbol{\phi_a}}(\boldsymbol{a}_t\mid \boldsymbol{s}_t)$ through the realized return $R(c;\boldsymbol{\zeta})$. 
As a result, the learning signal for the outer policy critically depends on the quality of the inner controller. Early in training, when $\pi_{\boldsymbol{\phi_a}}$ is still inaccurate and unstable, the estimated return can be noisy, potentially driving $\pi_{\boldsymbol{\phi}_c}$ toward suboptimal or misleading slider strategies. 

To address this dependency, we adopt a two-stage training procedure. 
In \textbf{Stage~1} (inner pretraining), the slider position is randomized each episode by sampling $c$ uniformly from $(c_{\min},c_{\max})$, while training only the inner-level SAC controller $\pi_{\boldsymbol{\phi_a}}$. 
This stage encourages the inner controller $\pi_{\boldsymbol{\phi_a}}$ to develop reliable thrust control across the full range of CoM configurations before introducing the outer-level decision making.

In \textbf{Stage~2} (joint training), both levels are trained simultaneously. The outer policy selects slider configuration $c\sim \pi_{\boldsymbol{\phi}_c}(c\mid \boldsymbol{\zeta})$ for the episode, and the inner controller is updated online under the resulting closed-loop dynamics. 
This stage enables coordinated exploration and refinement of both $\pi_{\boldsymbol{\phi}_c}$ and $\pi_{\boldsymbol{\phi_a}}$.

\begin{figure}[tbp]
\centerline{\includegraphics[width=1.0\linewidth]{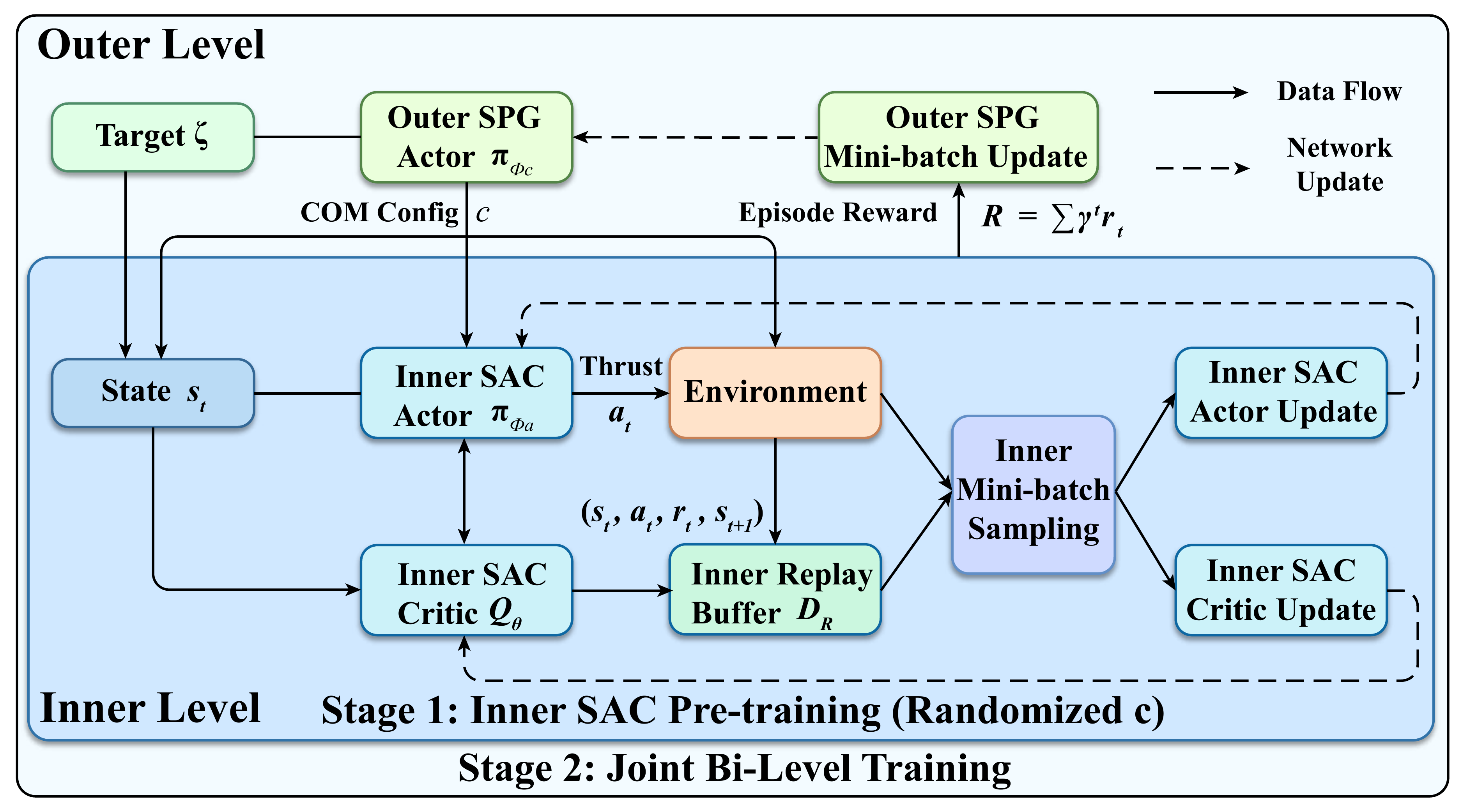}}
\caption{Overview of the proposed Bi-Level RL framework for RGBlimp. The outer level uses SPG to select the episode-wise slider position for CoM reconfiguration, while the inner level uses SAC to generate continuous thruster commands. Training proceeds in two stages. In Stage 1, the inner SAC controller is pretrained with randomly sampled slider positions in each episode. In Stage 2, the outer and inner policies are jointly optimized through bi-level training.}
\label{fig:control}
\end{figure}

Algorithm~\ref{algorithm} summarizes the complete bi-level RL pipeline, including the two-stage training schedule and the corresponding update rules for both levels. Fig.~\ref{fig:control} provides an intuitive illustration of the overall algorithmic framework.

\begin{algorithm}[tbp]
\caption{Two-Stage Bi-Level RL Training}
\label{algorithm}
\begin{algorithmic}[1]
\STATE \textbf{Initialize inner SAC:} critics $Q_{\boldsymbol{\theta}_1},Q_{\boldsymbol{\theta}_2}$, actor $\pi_{\boldsymbol{\phi}_a}$, temperature $\alpha$, target critics $Q_{\boldsymbol{\bar{\theta}}_1},Q_{\boldsymbol{\bar{\theta}}_2}$ with $\boldsymbol{\bar{\theta}}_i\gets\boldsymbol{\theta}_i$ for $i\in\{1,2\}$, and replay buffer $\mathcal{D}_R$.
\STATE \textbf{Initialize outer SPG:} policy $\pi_{\boldsymbol{\phi}_c}$ and temperature $\beta$.

\FOR{each episode $j=1$ \textbf{to} $M$}
    \STATE Sample target position $\boldsymbol{\zeta}\sim(\boldsymbol{\zeta}_{\min},\boldsymbol{\zeta}_{\max})$.
    \IF{$j<N$}
        \STATE Sample slider configuration $c\sim(c_{\min},c_{\max})$.
    \ELSE
        \STATE Sample $c \sim \pi_{\boldsymbol{\phi}_c}(c \mid \boldsymbol{\zeta})$.
    \ENDIF

    \FOR{each time step $t=0$ \textbf{to} $T-1$}
        \STATE Execute $\boldsymbol{a}_t \sim \pi_{\boldsymbol{\phi_a}}(\boldsymbol{a}_t \mid \boldsymbol{s}_t)$, observe $r_t,\boldsymbol{s}_{t+1}$, and store $(\boldsymbol{s}_t,\boldsymbol{a}_t,r_t,\boldsymbol{s}_{t+1})$ in $\mathcal{D}_R$.
        \STATE Sample minibatch $\mathcal{B}_a$ from $\mathcal{D}_R$ and compute $\nabla_{\boldsymbol{\theta}_i}\mathcal{L}_Q(\boldsymbol{\theta}_i)$, $\nabla_{\boldsymbol{\phi_a}}\mathcal{L}_{\pi}(\boldsymbol{\phi_a})$, and $\nabla_{\alpha}\mathcal{L}(\alpha)$.
        \STATE Update critics $\boldsymbol{\theta}_i \gets \boldsymbol{\theta}_i - \lambda_Q \nabla_{\boldsymbol{\theta}_i}\mathcal{L}_Q(\boldsymbol{\theta}_i)$ for $i\in\{1,2\}$, actor $\boldsymbol{\phi_a} \gets \boldsymbol{\phi_a} - \lambda_{\pi} \nabla_{\boldsymbol{\phi_a}}\mathcal{L}_{\pi}(\boldsymbol{\phi_a})$, and temperature $\alpha \gets \alpha - \lambda_{\alpha} \nabla_{\alpha}\mathcal{L}(\alpha)$.
        \STATE Update target critics $\boldsymbol{\bar{\theta}}_{i} \gets \rho\,\boldsymbol{\theta}_{i}+(1-\rho)\,\boldsymbol{\bar{\theta}}_{i}$ for $i\in\{1,2\}$.
    \ENDFOR

    \IF{$j\ge N$}
        \STATE Compute $R(c;\boldsymbol{\zeta})=\sum_{t=0}^{T}\gamma^t r(\boldsymbol{s}_t,\boldsymbol{a}_t)$ and add $(\boldsymbol{\zeta},c,R(c;\boldsymbol{\zeta}))$ to minibatch $\mathcal{B}_c$.
        \STATE Compute $\nabla_{\boldsymbol{\phi}_c} J_{\mathcal{O}}(\boldsymbol{\phi}_c)$ and $\nabla_{\beta}\mathcal{L}(\beta)$.
        \STATE Update outer policy $\boldsymbol{\phi}_c \gets \boldsymbol{\phi}_c + \eta \nabla_{\boldsymbol{\phi}_c} J_{\mathcal{O}}(\boldsymbol{\phi}_c)$ and temperature $\beta \gets \beta - \lambda_{\beta} \nabla_{\beta}\mathcal{L}(\beta)$.
    \ENDIF
\ENDFOR
\end{algorithmic}
\end{algorithm}

\subsection{Convergence Analysis}
We now analyze the convergence of the proposed bi-level RL algorithm. 
A key property is that the inner-level controller does not depend on the outer-level policy parameters. 
The inner policy $\pi_{\boldsymbol{\phi_a}}(\boldsymbol{a}_t\mid \boldsymbol{s}_t)$ is trained from environment interactions under the augmented state, which contains the episode-wise configuration $c$. 
Hence, convergence of the inner-level controller is treated following standard SAC results \cite{SAC-2018}. 

We therefore focus on the outer policy $\pi_{\boldsymbol{\phi}_c}(c\mid \boldsymbol{\zeta})$ and characterize how a non-converged inner controller affects the outer updates.
Let $k=0,1,2,\dots$ denote the outer-update iteration index in Stage~2, where $k=0$ is the first outer update after Stage~1 pretraining. 
Let $\pi_{\boldsymbol{\phi_a}}^{\,k}$ denote the inner policy at iteration $k$, and let $R^{k}(c;\boldsymbol{\zeta})$ denote the realized discounted return when executing $\pi_{\boldsymbol{\phi_a}}^{\,k}$ with the episode-wise configuration $c$.

Let $\pi_{\boldsymbol{\phi_a}}^*$ denote the converged inner policy for a fixed configuration $c$ with corresponding optimal discounted return
\begin{equation}
R^*(c;\boldsymbol{\zeta}) \triangleq 
\mathbb{E}_{\boldsymbol{\tau}\sim \pi^*_{\boldsymbol{\phi_a}}(\cdot \mid c,\boldsymbol{\zeta})}
\left[
\sum_{t=0}^{T}\gamma^t\, r(\boldsymbol{s}_t,\boldsymbol{a}_t)
\right],
\label{eq:outer_opt_return_def}
\end{equation}
where $t=0,\dots,T$ indexes time steps within an episode trajectory $\boldsymbol{\tau}$.

Denote the inner critic approximation error at iteration $k$  by nonnegative $\delta_Q^{\,k}$, and assume $\delta_Q^{\,k}\to 0$ as $k\to\infty$ \cite{zhang2020cofpac}. The induced return discrepancy is bounded by
\begin{equation}
\big|R^*(c;\boldsymbol{\zeta})-R^{k}(c;\boldsymbol{\zeta})\big|
\le
C_R(\gamma)\,\delta_Q^{\,k},
\label{eq:return_gap_bound_by_deltaQ}
\end{equation}
where $C_R(\gamma)>0$ depends on the discount factor and reward bounds.

The ideal outer objective is
\begin{equation}
J_{\mathcal{O}}^*(\boldsymbol{\phi}_c)\triangleq
\mathbb{E}_{\boldsymbol{\zeta}\sim \mathcal{D}}
\left[
\mathbb{E}_{c\sim \pi_{\boldsymbol{\phi}_c}(\cdot\mid \boldsymbol{\zeta})}
\left[
R^*(c;\boldsymbol{\zeta})
+\beta\,\mathcal{H}\!\left(\pi_{\boldsymbol{\phi}_c}(\cdot\mid \boldsymbol{\zeta})\right)
\right]
\right],
\label{eq:outer_obj_star}
\end{equation}
and the corresponding optimal outer-level parameters are
\begin{equation}
\boldsymbol{\phi}_c^* \triangleq \arg\max_{\boldsymbol{\phi}_c} J_{\mathcal{O}}^*(\boldsymbol{\phi}_c).
\label{eq:outer_opt_param}
\end{equation}

The gradient of the ideal objective is
\begin{equation*}
\small
\nabla_{\boldsymbol{\phi}_c} J_{\mathcal{O}}^*(\boldsymbol{\phi}_c)
=
\mathbb{E}_{\boldsymbol{\zeta}\sim \mathcal{D}}
\left[
\mathbb{E}_{c\sim \pi_{\boldsymbol{\phi}_c}(\cdot\mid \boldsymbol{\zeta})}
\left[
\nabla_{\boldsymbol{\phi}_c}\log \pi_{\boldsymbol{\phi}_c}(c\mid \boldsymbol{\zeta})\,R^*(c;\boldsymbol{\zeta})
\right.
\right.
\end{equation*}
\begin{equation}
\left.
\left.
+
\beta\,\nabla_{\boldsymbol{\phi}_c}\mathcal{H}\!\left(\pi_{\boldsymbol{\phi}_c}(\cdot\mid \boldsymbol{\zeta})\right)
\right]
\right],
\label{eq:outer_grad_star}
\end{equation}

In practice, the outer update at iteration $k$  uses rollouts generated by the current inner policy, yielding
\begin{equation}
J_{\mathcal{O}}^{k}(\boldsymbol{\phi}_c)\triangleq
\mathbb{E}_{\boldsymbol{\zeta}\sim \mathcal{D}}
\left[
\mathbb{E}_{c\sim \pi_{\boldsymbol{\phi}_c}(\cdot\mid \boldsymbol{\zeta})}
\left[
R^{k}(c;\boldsymbol{\zeta})
+\beta\,\mathcal{H}\!\left(\pi_{\boldsymbol{\phi}_c}(\cdot\mid \boldsymbol{\zeta})\right)
\right]
\right],
\label{eq:outer_obj_k}
\end{equation}
with gradient
\begin{equation*}
\small
\nabla_{\boldsymbol{\phi}_c} J_{\mathcal{O}}^{k}(\boldsymbol{\phi}_c)
=
\mathbb{E}_{\boldsymbol{\zeta}\sim \mathcal{D}}
\left[
\mathbb{E}_{c\sim \pi_{\boldsymbol{\phi}_c}(\cdot\mid \boldsymbol{\zeta})}
\left[
\nabla_{\boldsymbol{\phi}_c}\log \pi_{\boldsymbol{\phi}_c}(c\mid \boldsymbol{\zeta})\,R^{k}(c;\boldsymbol{\zeta})
\right.
\right.
\end{equation*}
\begin{equation}
\left.
\left.
+
\beta\,\nabla_{\boldsymbol{\phi}_c}\mathcal{H}\!\left(\pi_{\boldsymbol{\phi}_c}(\cdot\mid \boldsymbol{\zeta})\right)
\right]
\right].
\label{eq:outer_grad_k}
\end{equation}

We define the gradient bias at iteration $k$ as
\begin{equation}
\boldsymbol{\Delta}^{k} \triangleq 
\nabla_{\boldsymbol{\phi}_c} J_{\mathcal{O}}^*(\boldsymbol{\phi}_c)
-
\nabla_{\boldsymbol{\phi}_c} J_{\mathcal{O}}^{k}(\boldsymbol{\phi}_c).
\label{eq:grad_bias_def}
\end{equation}
Since the entropy-gradient terms cancel, the bias reduces to
\begin{equation*}
\small
\boldsymbol{\Delta}^{k}
=
\mathbb{E}_{\boldsymbol{\zeta}\sim \mathcal{D}}
\left[
\mathbb{E}_{c\sim \pi_{\boldsymbol{\phi}_c}(\cdot\mid \boldsymbol{\zeta})}
\left[
\nabla_{\boldsymbol{\phi}_c}\log \pi_{\boldsymbol{\phi}_c}(c\mid \boldsymbol{\zeta})
\left(R^*(c;\boldsymbol{\zeta})
\right.
\right.
\right.
\end{equation*}
\begin{equation}
\left.
\left.
\left.
-R^{k}(c;\boldsymbol{\zeta})\right)
\right]
\right].
\label{eq:grad_bias_expand}
\end{equation}

Assume the score function is uniformly bounded
\begin{equation}
\left\|\nabla_{\boldsymbol{\phi}_c}\log \pi_{\boldsymbol{\phi}_c}(c\mid \boldsymbol{\zeta})\right\|\le \varGamma .
\label{eq:score_bound}
\end{equation}
Then
\begin{equation}
\left\|\boldsymbol{\Delta}^{k}\right\|
\le
\varGamma\,
\mathbb{E}_{\boldsymbol{\zeta}\sim \mathcal{D}}
\left[
\mathbb{E}_{c\sim \pi_{\boldsymbol{\phi}_c}(\cdot\mid \boldsymbol{\zeta})}
\left[
\left|R^*(c;\boldsymbol{\zeta})-R^{k}(c;\boldsymbol{\zeta})\right|
\right]
\right],
\label{eq:grad_bias_bound_by_return_gap}
\end{equation}
and combining~\eqref{eq:grad_bias_bound_by_return_gap} with~\eqref{eq:return_gap_bound_by_deltaQ} yields
\begin{equation}
\left\|\boldsymbol{\Delta}^{k}\right\|
\le
\varGamma\,C_R(\gamma)\,\delta_Q^{\,k}.
\label{eq:grad_bias_bound_by_deltaQ}
\end{equation}

Stage~2 stochastic gradient ascent follows
\begin{equation*}
\boldsymbol{\phi}_c^{k+1}
=
\boldsymbol{\phi}_c^{k}
+
\eta^{k}\,
\widehat{\nabla}_{\boldsymbol{\phi}_c} J_{\mathcal{O}}^{k}\!\left(\boldsymbol{\phi}_c^{k}\right),
\end{equation*}
\begin{equation}
=
\boldsymbol{\phi}_c^{k}
+
\eta^{k}\,
\nabla_{\boldsymbol{\phi}_c} J_{\mathcal{O}}^{k}\!\left(\boldsymbol{\phi}_c^{k}\right)
+
\eta^{k}\,\boldsymbol{\xi}^{k},
\label{eq:outer_update_stochastic_stage2}
\end{equation}
where $\boldsymbol{\xi}^{k}$ is zero-mean noise with bounded second moment. Let $\mathcal{F}^{k}$ denote the training history up to iteration $k$, and $\sigma^{2}<\infty$ denote a uniform bound on the conditional second moment. Following conventional stochastic approximation and actor--critic analysis \cite{konda2003actor}, we assume
\begin{equation}
\mathbb{E}\!\left[\boldsymbol{\xi}^{k}\mid \mathcal{F}^{k}\right]=\boldsymbol{0}, 
\qquad 
\mathbb{E}\!\left[\|\boldsymbol{\xi}^{k}\|^2\mid \mathcal{F}^{k}\right]\le \sigma^2,
\label{eq:spg_noise_assumption}
\end{equation}
We also adopt the Robbins--Monro step-size conditions \cite{robbins1951stochastic}
\begin{equation}
\sum_{k=0}^{\infty}\eta^{k}=\infty,
\qquad
\sum_{k=0}^{\infty}(\eta^{k})^2<\infty.
\label{eq:robbins_monro}
\end{equation}

Using~\eqref{eq:grad_bias_def}, we have
\begin{equation}
\nabla_{\boldsymbol{\phi}_c} J_{\mathcal{O}}^{k}(\boldsymbol{\phi}_c)
=
\nabla_{\boldsymbol{\phi}_c} J_{\mathcal{O}}^{*}(\boldsymbol{\phi}_c)
-
\boldsymbol{\Delta}^{k}.
\label{eq:outer_grad_relation}
\end{equation}
Substituting~\eqref{eq:outer_grad_relation} into~\eqref{eq:outer_update_stochastic_stage2} gives
\begin{equation}
\boldsymbol{\phi}_c^{k+1}
=
\boldsymbol{\phi}_c^{k}
+
\eta^{k}\,
\nabla_{\boldsymbol{\phi}_c} J_{\mathcal{O}}^{*}\!\left(\boldsymbol{\phi}_c^{k}\right)
-
\eta^{k}\,\boldsymbol{\Delta}^{k}
+
\eta^{k}\,\boldsymbol{\xi}^{k}.
\label{eq:outer_update_biased}
\end{equation}

To ensure convergence of SPG on the ideal objective $J_{\mathcal{O}}^*$, it suffices that the gradient bias is asymptotically negligible in the stochastic approximation sense. A standard sufficient condition is
\begin{equation}
\sum_{k=0}^{\infty}\eta^{k}\left\|\boldsymbol{\Delta}^{k}\right\| < \infty,
\label{eq:bias_summability_condition}
\end{equation}
together with the standard regularity assumptions, such as Lipschitz continuity of $\nabla J_{\mathcal{O}}^*$ and bounded iterates \cite{konda2003actor}. Under~\eqref{eq:spg_noise_assumption}--\eqref{eq:robbins_monro} and~\eqref{eq:bias_summability_condition}, the sequence $\{\boldsymbol{\phi}_c^{k}\}$ converges almost surely to the set of stationary points of $J_{\mathcal{O}}^*$.

A sufficient condition for~\eqref{eq:bias_summability_condition} follows from~\eqref{eq:grad_bias_bound_by_deltaQ}. Specifically, if
\begin{equation}
\delta_Q^{\,k}=\mathcal{O}\!\left(\frac{1}{\sqrt{k+1}}\right),
\label{eq:delta_rate_assumption}
\end{equation}
and
\begin{equation}
\eta^{k}=\mathcal{O}\!\left(\frac{1}{k+1}\right),
\label{eq:eta_rate_assumption}
\end{equation}
then $\eta^{k}\left\|\boldsymbol{\Delta}^{k}\right\|=\mathcal{O}((k+1)^{-3/2})$ and~\eqref{eq:bias_summability_condition} holds.

Stage~1 pretraining ensures that the inner controller is close to convergence at the start of Stage~2, so $\delta_Q^{\,0}$ and $\|\boldsymbol{\Delta}^{0}\|$ are small. This mitigates early perturbations in the outer updates and improves practical stability. Moreover, if $J_{\mathcal{O}}^{*}$ is concave or admits a unique maximizer, the stationary point coincides with the global optimum $\boldsymbol{\phi}_c^{*}$ in~\eqref{eq:outer_opt_param}, i.e.,
\begin{equation}
\lim_{k\to\infty}\boldsymbol{\phi}_c^{k}=\boldsymbol{\phi}_c^{*}.
\label{eq:outer_converges_to_opt}
\end{equation}

\section{Training}
\subsection{Training Setup}
\subsubsection{Domain Randomization}
To reduce the sim-to-real gap, we employ domain randomization during training. At the beginning of each episode, the initial state is perturbed by small random offsets to mimic variations in launch conditions and state estimation errors. In addition, the aerodynamic parameters are uniformly randomized within $\pm 10\%$ of their nominal identified values, whereas the mass, buoyancy, and inertia terms are randomized within $\pm 5\%$. This setting captures moderate parameter uncertainty and unmodeled effects in the simulator.

\subsubsection{Parameter Setup}
During the training process, the target position $\boldsymbol{\zeta}$ is sampled uniformly from a 3D workspace defined by
$x\in[4,5]\,\mathrm{m}$, $y\in[-2,2]\,\mathrm{m}$, and $z\in[-1,1]\,\mathrm{m}$. 
Each episode lasts up to $15\,\mathrm{s}$ and terminates early when the blimp reaches a goal region defined as a ball of radius $0.2\,\mathrm{m}$ centered at $\boldsymbol{\zeta}$. 
The slider position is constrained to $[-5,\,5]\,\mathrm{cm}$. 
The Stage 1 pretraining comprises $N=15000$ episodes, and the total number of training episodes is $M=30000$. 
The SAC critic and actor are implemented as MLPs with hidden sizes 512 and 128, respectively.

\subsubsection{Reward Function}
We use a shaped reward to encourage accurate and efficient goal-directed tracking control. 
At each time step, the reward is designed as
\begin{equation}
r_t = -2.0\, e_{\text{trk},t} - e_{\text{head},t} + 2.0\, \Delta d_t + r_{\text{target},t},
\label{eq:reward_def}
\end{equation}
where $e_{\text{trk},t}$ is the shortest distance from the current position $\boldsymbol{p}_t$ to the straight-line path from the origin to the target $\boldsymbol{\zeta}$, $e_{\text{head},t}$ is the heading error between the velocity direction and the direction to the target $\boldsymbol{\zeta}$, and $\Delta d_t$ denotes the forward progress between two consecutive time steps
\begin{equation}
\Delta d_t \triangleq \|\boldsymbol{p}_{t-1}-\boldsymbol{\zeta}\|-\|\boldsymbol{p}_{t}-\boldsymbol{\zeta}\|.
\label{eq:progress_def}
\end{equation}
A terminal bonus is applied upon reaching the goal
\begin{equation}
r_{\text{target},t}=
\begin{cases}
20, & \|\boldsymbol{p}_t-\boldsymbol{\zeta}\|\le r_g,\\
0, & \text{otherwise},
\end{cases}
\label{eq:reward_target}
\end{equation}
Reward weights are empirically tuned to balance tracking accuracy, heading alignment, and forward progress.

\subsection{Training Results}
\begin{figure}[tbp]
\centerline{\includegraphics[width=1.0\linewidth]{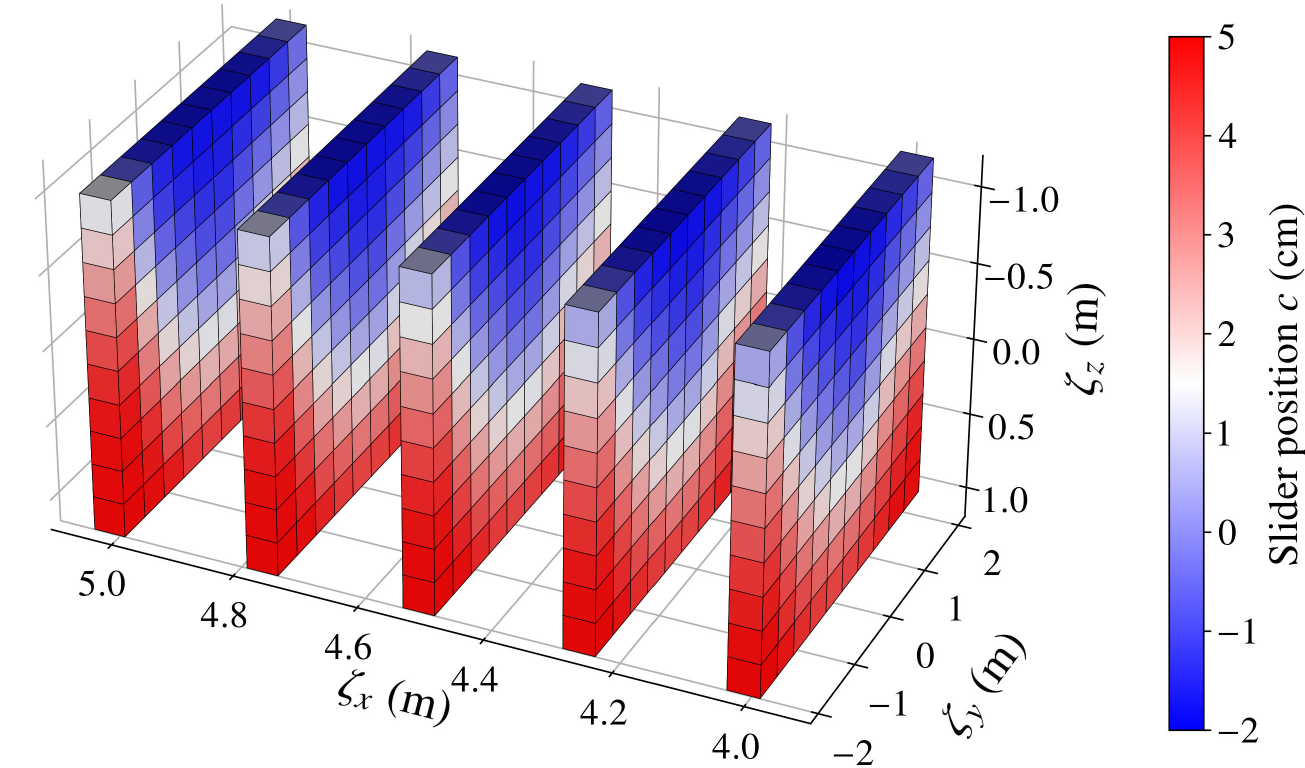}}
\caption{Learned outer-level slider policy $\pi_{\boldsymbol{\phi}_c}(c\mid \boldsymbol{\zeta})$. Each voxel represents a target $\boldsymbol{\zeta}=[\zeta_x,\zeta_y,\zeta_z]^\top$, with color indicating the selected slider configuration $c$ (cm). The policy shows strong symmetry with respect to $\zeta_y$ and varies primarily with target height, moving the slider backward for higher targets and forward for lower targets.}
\label{fig:outer_slider_policy}
\end{figure}

The proposed bi-level training procedure converges reliably in simulation. Fig.~\ref{fig:outer_slider_policy} visualizes the learned outer-loop slider policy $\pi_{\boldsymbol{\phi}_c}(c\mid \boldsymbol{\zeta})$ over different target locations.

Each voxel corresponds to a target $\boldsymbol{\zeta}$, and the color encodes the episode-wise slider configuration $c$ in cm. Two main trends are observed. First, although the slider is allowed to move within $[-5,\,5]$\,cm, the learned policy concentrates on a narrower interval of roughly $[-2,\,5]$\,cm, reflecting that $c=0$ denotes the mechanical zero of the slider rather than a physically critical CoM. Second, the policy aligns with physical intuition. For higher targets (smaller $\zeta_z$), the slider moves backward to promote nose-up pitch and facilitate climbing; for lower targets, it moves forward to induce nose-down pitch and assist descending.

The learned map exhibits strong left--right symmetry about $\zeta_y=0$, indicating consistent lateral behavior. Along the $\zeta_x$ direction, the dependence is weaker but still visible. Closer targets require slightly more backward slider positions to achieve the same altitude change. Overall, the learned outer-loop strategy is coherent and physically interpretable.

\section{Experiment}

\subsection{Experimental Setup}
Experiments are conducted in a $6\,\mathrm{m}\times 4\,\mathrm{m}\times 3\,\mathrm{m}$ motion-capture arena equipped with 20 OptiTrack cameras, streaming at $60\,\mathrm{Hz}$ and providing position estimates with $0.76\,\mathrm{mm}$ RMS accuracy. Twelve active markers are mounted on the blimp envelope to provide  state measurements for feedback control.

The goal-directed tracking control task consists of 27 target positions arranged on a $3\times 3\times 3$ grid:
$\zeta_x\in\{4.0,\,4.5,\,5.0\}\,\mathrm{m}$, $\zeta_y\in\{-2.0,\,0,\,2.0\}\,\mathrm{m}$, and $\zeta_z\in\{-1.0,\,0,\,1.0\}\,\mathrm{m}$.
All trials start from $\boldsymbol{p}_0=[0,0,0]^\top$ with the blimp facing the positive $x$ direction. This setting covers climbing, level-flight, and descending motions, as well as lateral and forward maneuvers. Each target is tested in three repeated trials under nominal initial conditions.

Five controllers are evaluated. The proposed method, \textbf{Bi-Level RL}, uses the learned outer policy $\pi_{\boldsymbol{\phi}_c}(c\mid \boldsymbol{\zeta})$ and  inner policy $\pi_{\boldsymbol{\phi_a}}(\boldsymbol{a}_t\mid \boldsymbol{s}_t)$. To isolate the effect of CoM reconfiguration, three baselines use the same learned inner SAC thrust controller with fixed slider positions, denoted as \textbf{SAC-Fixed ($-5$)}, \textbf{SAC-Fixed ($0$)}, and \textbf{SAC-Fixed ($5$)}, where the number represents $c\in\{-5,0,5\}\,\mathrm{cm}$.

We further evaluate the inner-level controller by replacing the learned SAC with a conventional PID while retaining the learned outer policy, denoted as \textbf{PID-SPG}. At each time step, the PID computes desired attitude angles from the current position $\boldsymbol{p}_t$ and the target $\boldsymbol{\zeta}$. Let $\boldsymbol{e}_{p,t}=\boldsymbol{\zeta}-\boldsymbol{p}_t$ and $\hat{\boldsymbol{e}}_{p,t}=\boldsymbol{e}_{p,t}/\|\boldsymbol{e}_{p,t}\|$. The yaw and pitch references are
\begin{equation}
\psi^{\mathrm{ref}}_t=\mathrm{atan2}(\hat{e}_{p,t,y},\,\hat{e}_{p,t,x}),
\label{eq:pid_yaw_ref}
\end{equation}
\begin{equation}
\theta^{\mathrm{ref}}_t=\mathrm{atan2}\!\Big(-\hat{e}_{p,t,z},\,\sqrt{\hat{e}_{p,t,x}^2+\hat{e}_{p,t,y}^2}\Big).
\label{eq:pid_pitch_ref}
\end{equation}
Two PID loops then generate the pitch and yaw moment commands $(\tau_{\theta,t},\,\tau_{\psi,t})$, which are mapped to the left and right thruster forces via the previously defined allocation matrix $\boldsymbol{B}\in\mathbb{R}^{2\times 6}$. Let $\boldsymbol{B}_{\theta\psi}\in\mathbb{R}^{2\times 2}$ denote the submatrix formed by the pitch and yaw columns of $\boldsymbol{B}$. Then
\begin{equation}
\begin{bmatrix} f_{l,t} \\ f_{r,t} \end{bmatrix}
=
\boldsymbol{B}_{\theta\psi}
\begin{bmatrix} \tau_{\theta,t} \\ \tau_{\psi,t} \end{bmatrix}
+
f_{\mathrm{ff}}(\boldsymbol{\zeta})
\begin{bmatrix} 1 \\ 1 \end{bmatrix},
\label{eq:pid_allocation}
\end{equation}
where $f_{\mathrm{ff}}(\boldsymbol{\zeta})$ is a goal-dependent feedforward term tuned to ensure sufficient forward motion. Owing to the severe underactuation and strong coupling of the blimp dynamics, designing an effective PID baseline requires careful heuristics and is less straightforward than learning-based control.

In the following, we first assess the outer-level CoM reconfiguration strategy and then compare the inner-level thrust controller against the PID baseline.

\begin{figure*}[tbp]
\centerline{\includegraphics[width=1.0\linewidth]{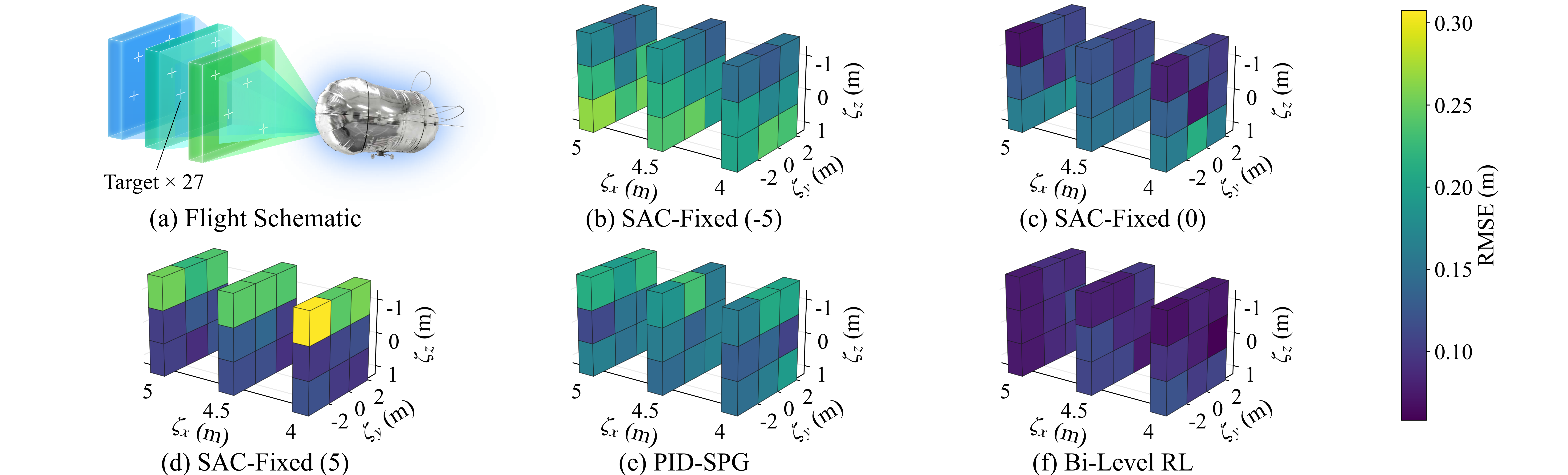}}
\caption{3D cross-track RMSE for flights from the origin to 27 target points. (a) Task setup of goal-directed tracking from a common start position to different targets. (b)--(f) RMSE distributions of different controllers over the $3\times3\times3$ target grid. Each voxel represents the mean RMSE over repeated trials for one target.}
\label{fig:outer_rmse_voxel}
\end{figure*}

\begin{figure*}[!t]
  \centering
    \subfloat[$\boldsymbol{\zeta}=(4.5,\,-2.0,\,-1.0)\,\mathrm{m}$]{\includegraphics[width=0.328\linewidth]{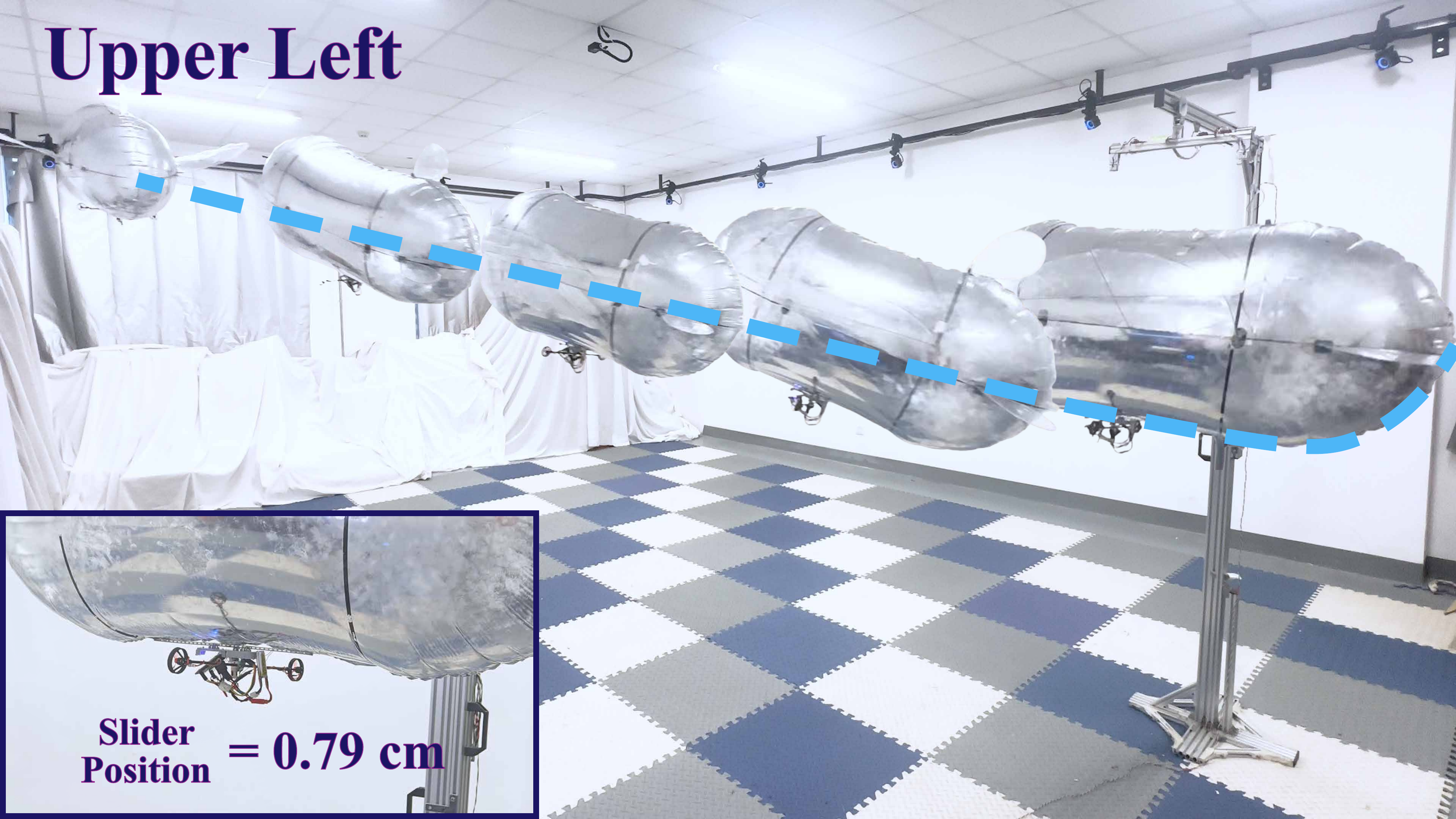}\label{fig:t0}}
    \hfil
    \subfloat[$\boldsymbol{\zeta}=(4.5,\,0.0,\,0.0)\,\mathrm{m}$]{\includegraphics[width=0.328\linewidth]{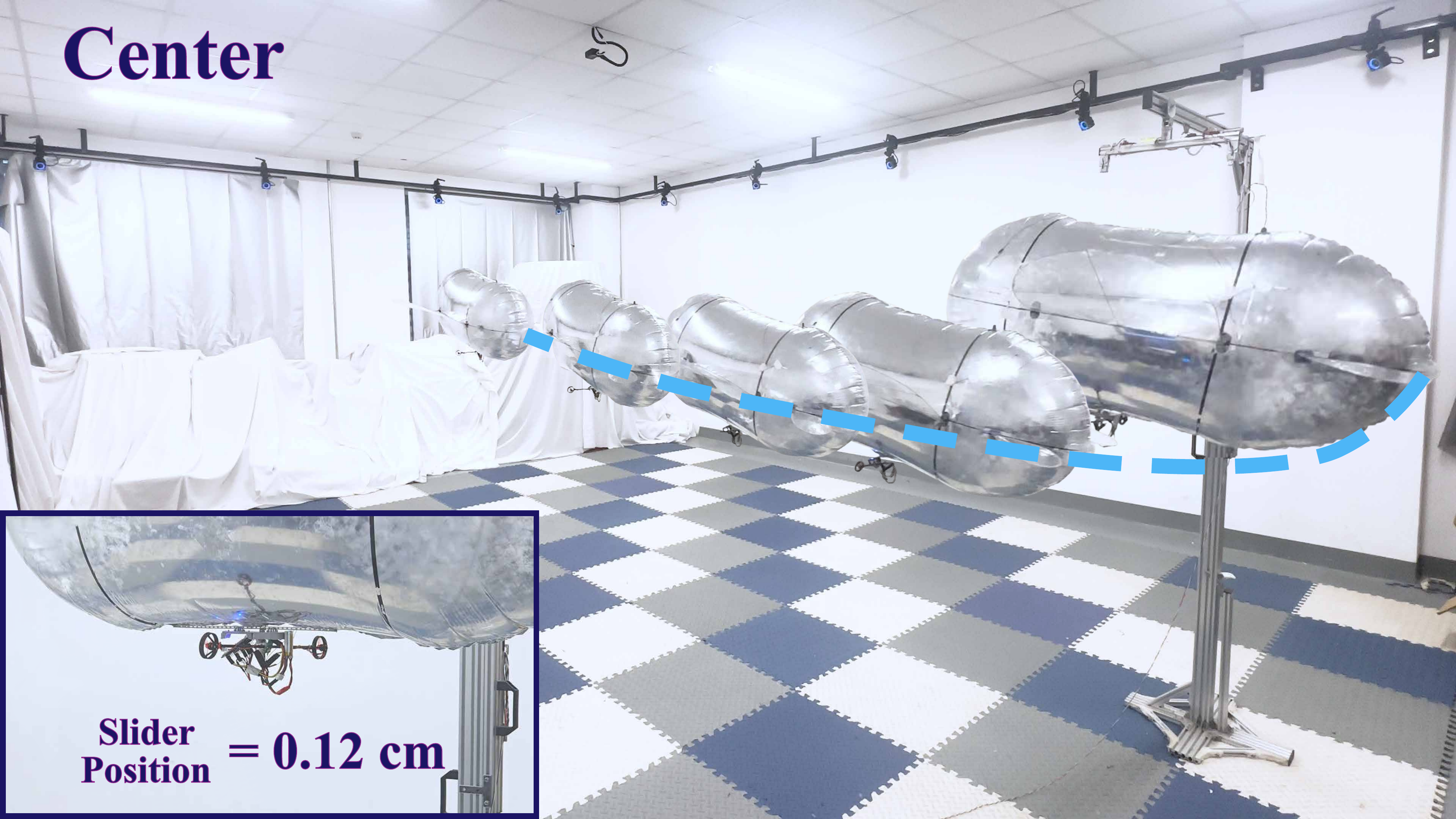}\label{fig:t0}}
    \hfil
    \subfloat[$\boldsymbol{\zeta}=(4.5,\,2.0,\,1.0)\,\mathrm{m}$]{\includegraphics[width=0.328\linewidth]{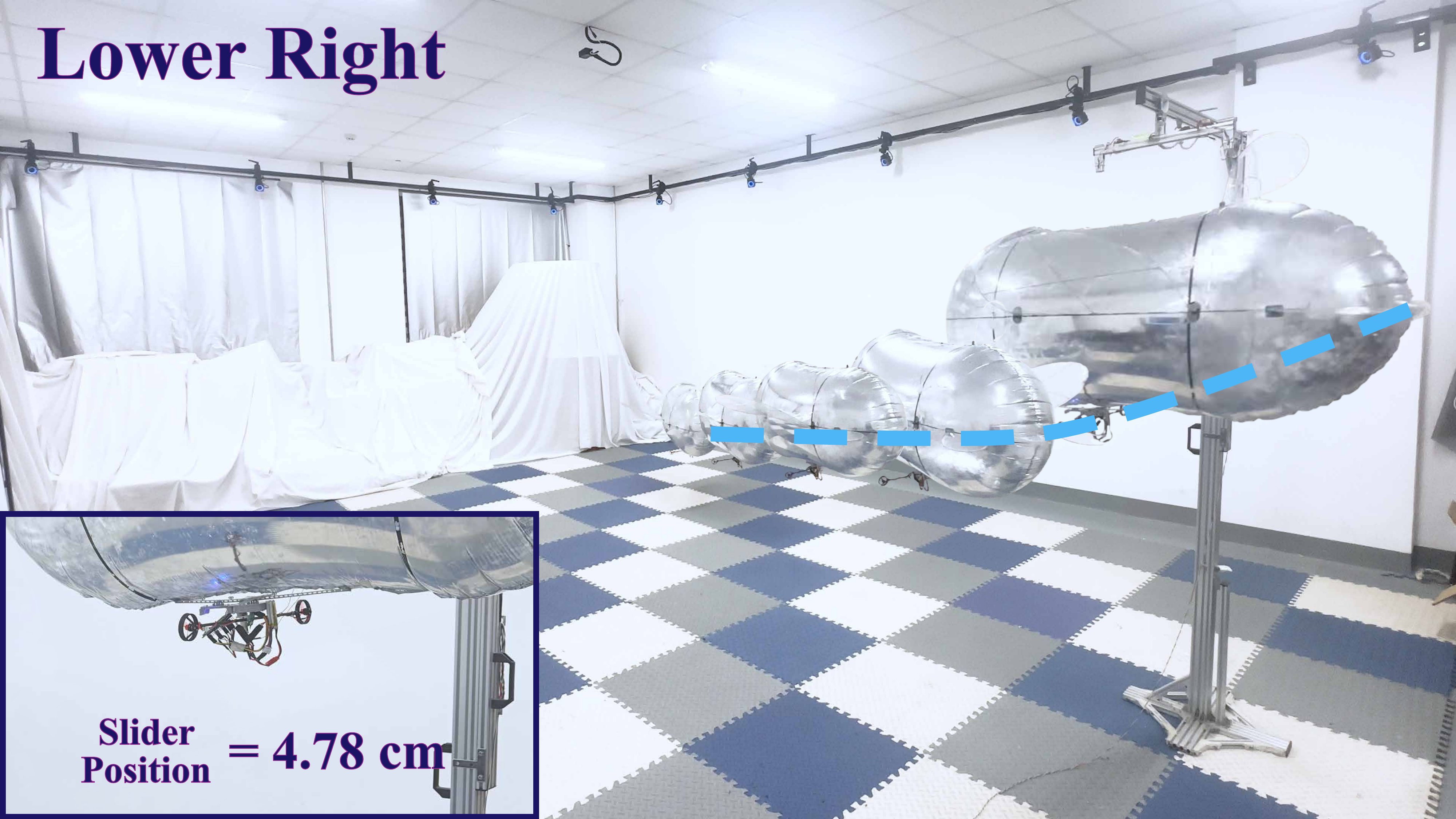}\label{fig:t0}}
    \caption{Flight snapshots of Bi-Level RL for three representative targets. The slider position is adaptively adjusted according to the target, after which the blimp follows the reference path and reaches the goal.}
  \label{fig:snapshots}
\end{figure*}

\begin{figure*}[tbp]
\centerline{\includegraphics[width=1.0\linewidth]{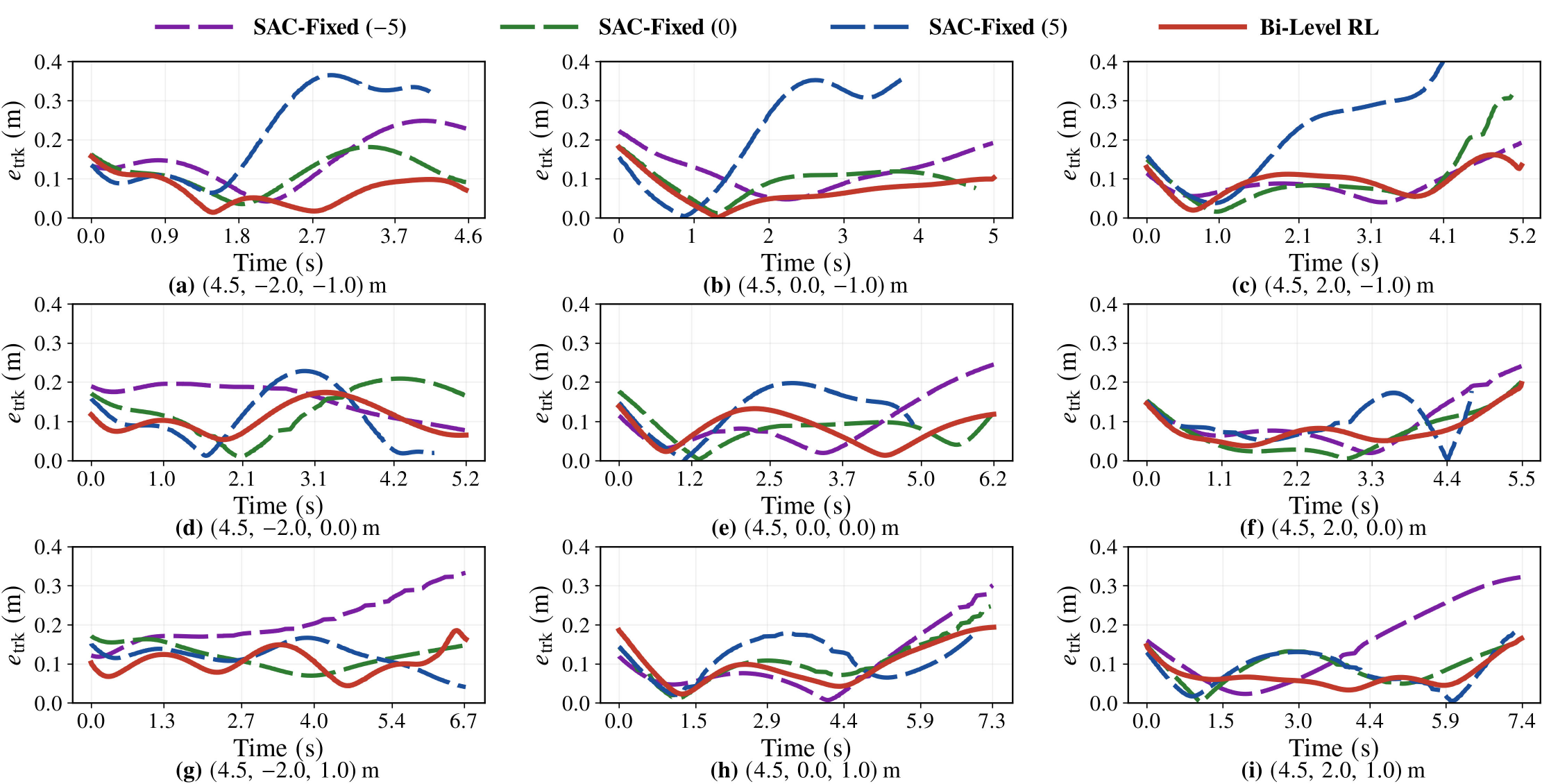}}
\caption{Time histories of cross-track error $e_{\mathrm{trk}}$ for nine representative targets on the $\zeta_x=4.5\,\mathrm{m}$ slice. (a)--(i) correspond to different combinations of $\zeta_y\in\{-2,0,2\}\,\mathrm{m}$ and $\zeta_z\in\{-1,0,1\}\,\mathrm{m}$. The dashed curves denote the fixed-slider baselines SAC-Fixed ($-5$), SAC-Fixed ($0$), and SAC-Fixed ($5$), while the solid curve denotes Bi-Level RL.}
\label{fig:outer_time_curve}
\end{figure*}

\subsection{Outer-Level Slider Policy Evaluation}

We first evaluate the learned outer-level slider strategy by comparing Bi-Level RL against three fixed-slider baselines, all using the same trained inner SAC controller. This comparison isolates the effect of target-dependent slider adaptation on tracking performance.

Fig.~\ref{fig:outer_rmse_voxel} shows RMSE distributions across the $3\times 3\times 3$ target grid. Tracking performance is strongly correlated with target height $\zeta_z$. The fixed-slider controllers exhibit a clear height-dependent trend. SAC-Fixed ($-5$) performs well for climbing targets but degrades for level and descending targets. SAC-Fixed ($0$) achieves moderate performance but worsens for descending targets. SAC-Fixed ($5$) performs well for level and descending targets but becomes much worse for climbing targets. For example, at the climbing target $\boldsymbol{\zeta}=(4.0,\,-2.0,\,-1.0)\,\mathrm{m}$, the RMSE of SAC-Fixed ($5$) reaches $0.308\,\mathrm{m}$. These trends align with physical intuition. A rearward slider favors nose-up motion and climbing, while a forward slider favors nose-down motion and descending. Hence, no single fixed slider  performs consistently across all tasks.

Table~\ref{tab-RMSE} quantifies this observation. For climbing targets with $\zeta_z=-1$, the mean RMSE values are $0.150\,\mathrm{m}$ for SAC-Fixed ($-5$), $0.103\,\mathrm{m}$ for SAC-Fixed ($0$), and $0.249\,\mathrm{m}$ for SAC-Fixed ($5$). For descending targets with $\zeta_z=1$, the ordering reverses, and SAC-Fixed ($5$) achieves the best $0.106\,\mathrm{m}$ RMSE, compared with $0.166\,\mathrm{m}$ for SAC-Fixed ($0$) and $0.234\,\mathrm{m}$ for SAC-Fixed ($-5$). These results confirm that the optimal fixed-slider placement depends on target height, motivating task-dependent outer-loop adaptation.

Bi-Level RL achieves the lowest RMSE in all height groups. Its mean RMSE is $0.082\,\mathrm{m}$ for climbing, $0.088\,\mathrm{m}$ for level flight, and $0.098\,\mathrm{m}$ for descending, with an overall average of $0.089\,\mathrm{m}$ over all 27 targets. As illustrated in Fig.~\ref{fig:outer_rmse_voxel}, the learned policy exhibits no region-specific weaknesses, maintaining uniformly low errors. Beyond target height, it fine-tunes slider positions for different $\zeta_x$ and $\zeta_y$, further improving tracking accuracy.

Representative flight snapshots are illustrated in Fig.~\ref{fig:snapshots}. The slider is adjusted per target, after which the blimp follows the reference path with minimal deviation, demonstrating effective task-dependent CoM configuration.

Fig.~\ref{fig:outer_time_curve} presents the time histories of the cross-track error $e_{\mathrm{trk}}$ for nine targets on the $\zeta_x=4.5\,\mathrm{m}$ slice. Bi-Level RL maintains errors below $0.1\,\mathrm{m}$ for most of the flight. In comparison, fixed-slider baselines show larger errors in their non-preferred regimes. SAC-Fixed ($-5$) struggles with descending targets, with final error exceeding $0.3\,\mathrm{m}$, and SAC-Fixed ($5$) performs poorly for climbing targets, where the error approaches up to $0.4\,\mathrm{m}$. SAC-Fixed ($0$) is more stable but still less consistent than Bi-Level RL.

Overall, these results demonstrate that slider configuration is highly task-dependent, with target height as the dominant factor. Bi-Level RL consistently selects suitable positions, achieving lower tracking errors across the full 27-goal evaluation set.

\begin{table}[tb]
\renewcommand{\arraystretch}{1.25}
\setlength{\tabcolsep}{4.6pt}
\centering
\caption{RMSE of 3D tracking error grouped by target height $\zeta_z$. Each entry reports mean $\pm$ standard deviation over all targets in the corresponding height group, in units of $\times 10^{-1}\,\mathrm{m}$. The three groups correspond to $\zeta_z\in\{-1,0,1\}\,\mathrm{m}$, and ``Overall'' summarizes all 27 targets.}
\label{tab-RMSE}
\begin{tabular}{@{}lcccc@{}}
\toprule
\multirow{2}{*}{\textbf{Controller}} &
\multicolumn{4}{c}{\textbf{Target height group} ($\times 10^{-1}\,\mathrm{m}$)} \\
\cmidrule(l){2-5}
& \makecell{\textbf{Climb}\\[-1pt]\small($\zeta_z{=}{-}1$)}
& \makecell{\textbf{Level}\\[-1pt]\small($\zeta_z{=}0$)}
& \makecell{\textbf{Descent}\\[-1pt]\small($\zeta_z{=}1$)}
& \makecell{\textbf{Overall}\\[-1pt]\small(all targets)} \\ \midrule
SAC-Fixed (-5) & 1.50 $\pm$ 0.20 & 1.94 $\pm$ 0.23 & 2.34 $\pm$ 0.24 & 1.93 $\pm$ 0.41 \\
SAC-Fixed (0) & 1.03 $\pm$ 0.18 & 1.16 $\pm$ 0.24 & 1.66 $\pm$ 0.22 & 1.28 $\pm$ 0.34 \\
SAC-Fixed (5) & 2.49 $\pm$ 0.24 & 1.16 $\pm$ 0.14 & 1.06 $\pm$ 0.12 & 1.57 $\pm$ 0.69 \\
PID-SPG & 1.97 $\pm$ 0.36 & 1.38 $\pm$ 0.29 & 1.60 $\pm$ 0.28 & 1.65 $\pm$ 0.47 \\
Bi-Level RL & 0.82 $\pm$ 0.09 & 0.88 $\pm$ 0.17 & 0.98 $\pm$ 0.16 & 0.89 $\pm$ 0.16 \\
\bottomrule
\end{tabular}
\end{table}

\begin{figure*}[tbp]
\centerline{\includegraphics[width=1.0\linewidth]{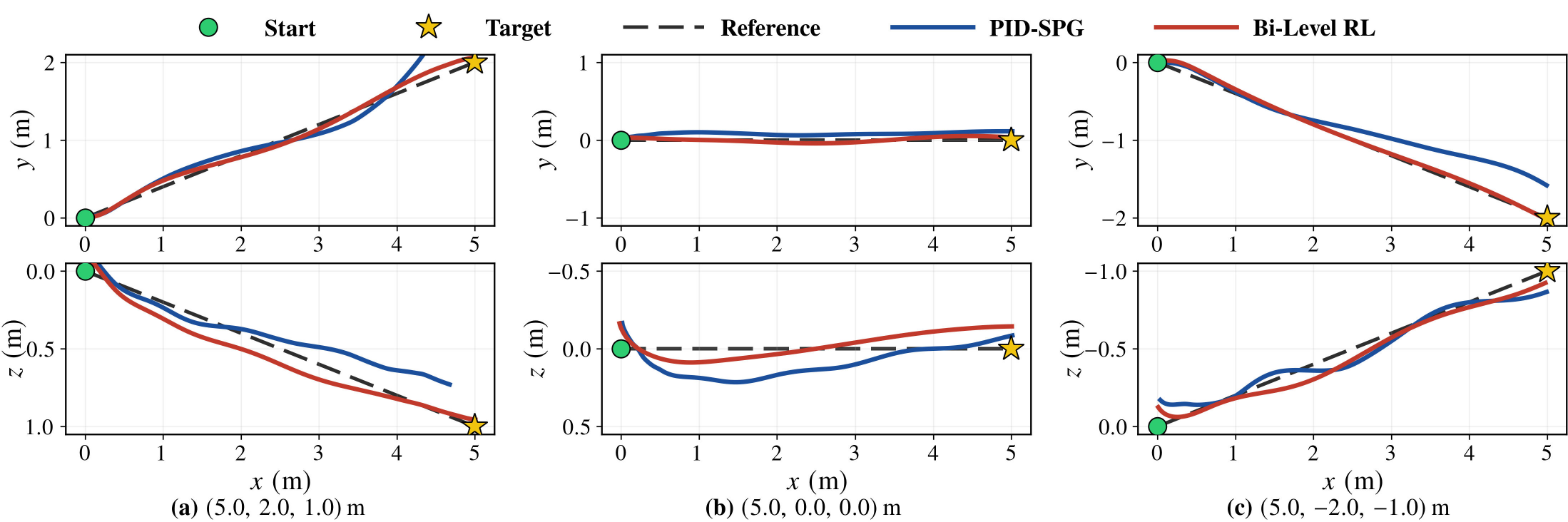}}
\caption{Comparison of trajectories between the learned inner SAC controller (Bi-Level RL) and the PID inner controller with the learned slider policy (PID-SPG). 
Top and bottom rows show  $x$--$y$ and $x$--$z$ plane projections, respectively. 
Columns (a)--(c) correspond to targets $\boldsymbol{\zeta}=(5.0,\,2.0,\,1.0)\,\mathrm{m}$, $(5.0,\,0.0,\,0.0)\,\mathrm{m}$, and $(5.0,\,-2.0,\,-1.0)\,\mathrm{m}$. 
Dashed lines denote the straight-line references from the start to the target.}
\label{fig:traj_pid_vs_birl}
\end{figure*}

\subsection{Inner-Level Controller Evaluation}
We evaluate the inner-level controller by comparing the learned SAC controller in Bi-Level RL with PID-SPG. Three representative targets are considered, namely $\boldsymbol{\zeta}=(5.0,\,2.0,\,1.0)\,\mathrm{m}$, $(5.0,\,0.0,\,0.0)\,\mathrm{m}$, and $(5.0,\,-2.0,\,-1.0)\,\mathrm{m}$. Fig.~\ref{fig:traj_pid_vs_birl} illustrates the corresponding trajectory projections on the $x$--$y$ and $x$--$z$ planes.

For $\boldsymbol{\zeta}=(5.0,\,2.0,\,1.0)\,\mathrm{m}$, Bi-Level RL closely follows the straight-line reference, whereas PID-SPG deviates in both $y$ and $z$ after $x$ exceeds $4\,\mathrm{m}$. For $(5.0,\,0.0,\,0.0)\,\mathrm{m}$, Bi-Level RL again tracks more tightly with a mean lateral deviation of approximately  $0.03\,\mathrm{m}$, compared to over $0.08\,\mathrm{m}$ for PID-SPG. In the $x$--$z$ plane, PID-SPG drops more than $0.2\,\mathrm{m}$ below the reference around $x\approx 1.5\,\mathrm{m}$. For $(5.0,\,-2.0,\,-1.0)\,\mathrm{m}$, PID-SPG accumulates over $0.4\,\mathrm{m}$ lateral deviation and exhibits noticeable vertical oscillation in the $x$--$z$ plane. By contrast, Bi-Level RL maintains  mean tracking errors of about $0.05\,\mathrm{m}$ in both projections.

The same trend is observed in the RMSE voxel plots in Fig.~\ref{fig:outer_rmse_voxel}. Over all 27 targets, PID-SPG yields consistently larger errors than Bi-Level RL. Although the outer-level slider policy still adapts the CoM configuration, the manually tuned PID inner loop struggles with the blimp's strong coupling and severe underactuation, leading to systematically larger tracking errors. Table~\ref{tab-RMSE} further supports this result. PID-SPG exhibits an overall standard deviation of $0.047\,\mathrm{m}$, roughly three times that of Bi-Level RL at $0.016\,\mathrm{m}$, indicating greater sensitivity to initial conditions and disturbances. Overall, the learned SAC inner policy provides superior robustness and more consistent flight performance.

\subsection{Overall Discussion}
The experimental results demonstrate the effectiveness of the proposed bi-level reinforcement learning framework. As summarized in Table~\ref{tab-RMSE}, Bi-Level RL achieves the lowest overall 3D tracking RMSE among all compared methods. Relative to SAC-Fixed ($-5$), SAC-Fixed ($0$), and SAC-Fixed ($5$), the total RMSE is reduced by $53.9\%$, $30.5\%$, and $43.3\%$, respectively, and by $46.1\%$ compared with PID-SPG. These results indicate that neither a fixed slider configuration nor a manually tuned inner-loop controller is sufficient for the complete goal-directed tracking control task. The best performance is obtained only when the outer loop adaptively configures the CoM according to the target and the inner loop simultaneously learns a thrust control strategy suited to the highly underactuated and strongly coupled blimp dynamics. Overall, the proposed bi-level design improves tracking accuracy, robustness, and consistency across the entire 27-goal evaluation set.

\section{Conclusion}
This paper presented a bi-level RL framework for goal-directed tracking control of an underactuated blimp with CoM reconfiguration. The proposed method decouples task-level slider selection from continuous thrust control, and combines a two-stage training strategy with convergence guarantees to improve learning stability. Experimental results in real flight demonstrated that the proposed framework consistently outperformed fixed-slider and PID-based baselines, confirming the effectiveness of integrating CoM adaptation with learned inner-loop control for this highly underactuated system.

Future work will focus on improving disturbance rejection under wind and model uncertainty, and extending the proposed framework to more challenging outdoor autonomous flight scenarios.

\bibliographystyle{IEEEtran}
\bibliography{IEEEabrv,paper}

\raggedbottom 

\begin{IEEEbiography}[{\includegraphics[width=1in,height=1.25in,clip,keepaspectratio]{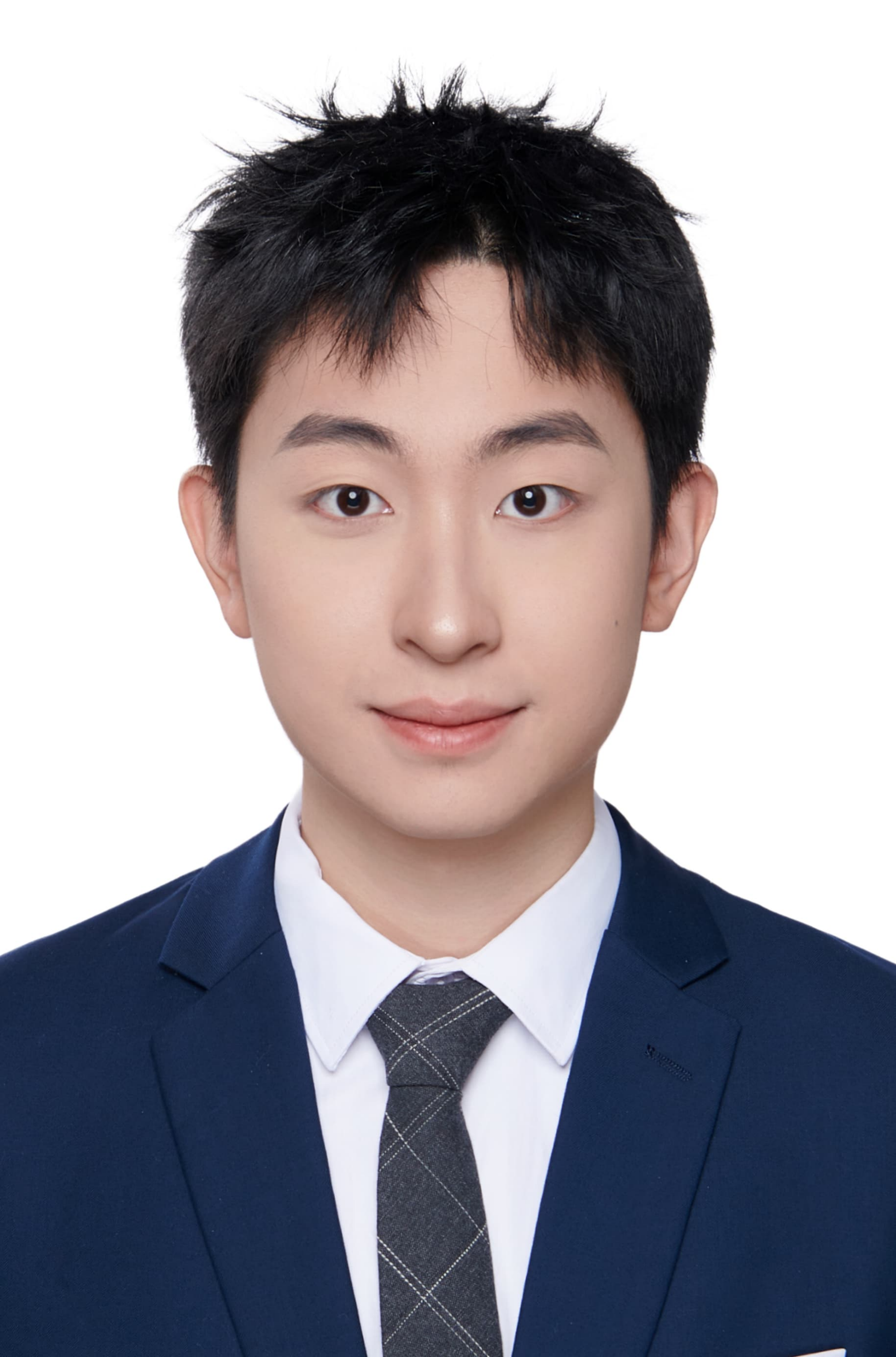}}]{Xiaorui Wang} (Student Member, IEEE) 
received the bachelor's degree in robotics engineering in 2025 from Peking University, Beijing, China, where he is currently working toward the Ph.D. degree in general mechanics and foundation of mechanics with the School of Advanced Manufacturing and Robotics, Peking University, Beijing, China. 

His research interests include aerial vehicles, dynamical modeling, and learning-based control.
\end{IEEEbiography}

\begin{IEEEbiography}[{\includegraphics[width=1in,height=1.25in,clip,keepaspectratio]{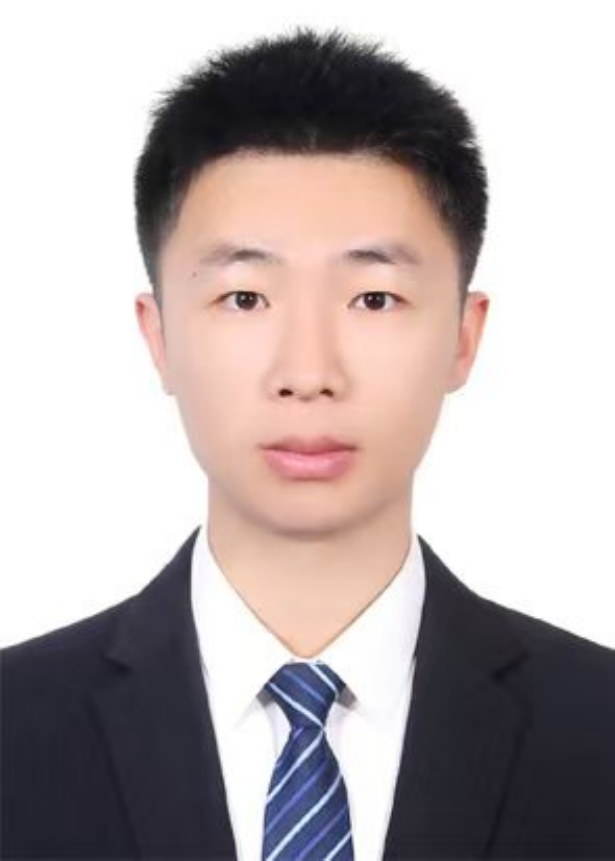}}]{Hongwu Wang} 
received the bachelor's degree in Robotics Engineering from Harbin Institute of Technology, Harbin, China, in 2024. He is currently working toward the M.S. degree in mechanical engineering with the School of Advanced Manufacturing and Robotics, Peking University, Beijing, China.  

His research interests include aerial vehicles, mechatronics
systems, reinforcement learning and robotic control.
\end{IEEEbiography}

\begin{IEEEbiography}[{\includegraphics[width=1in,height=1.25in,clip,keepaspectratio]{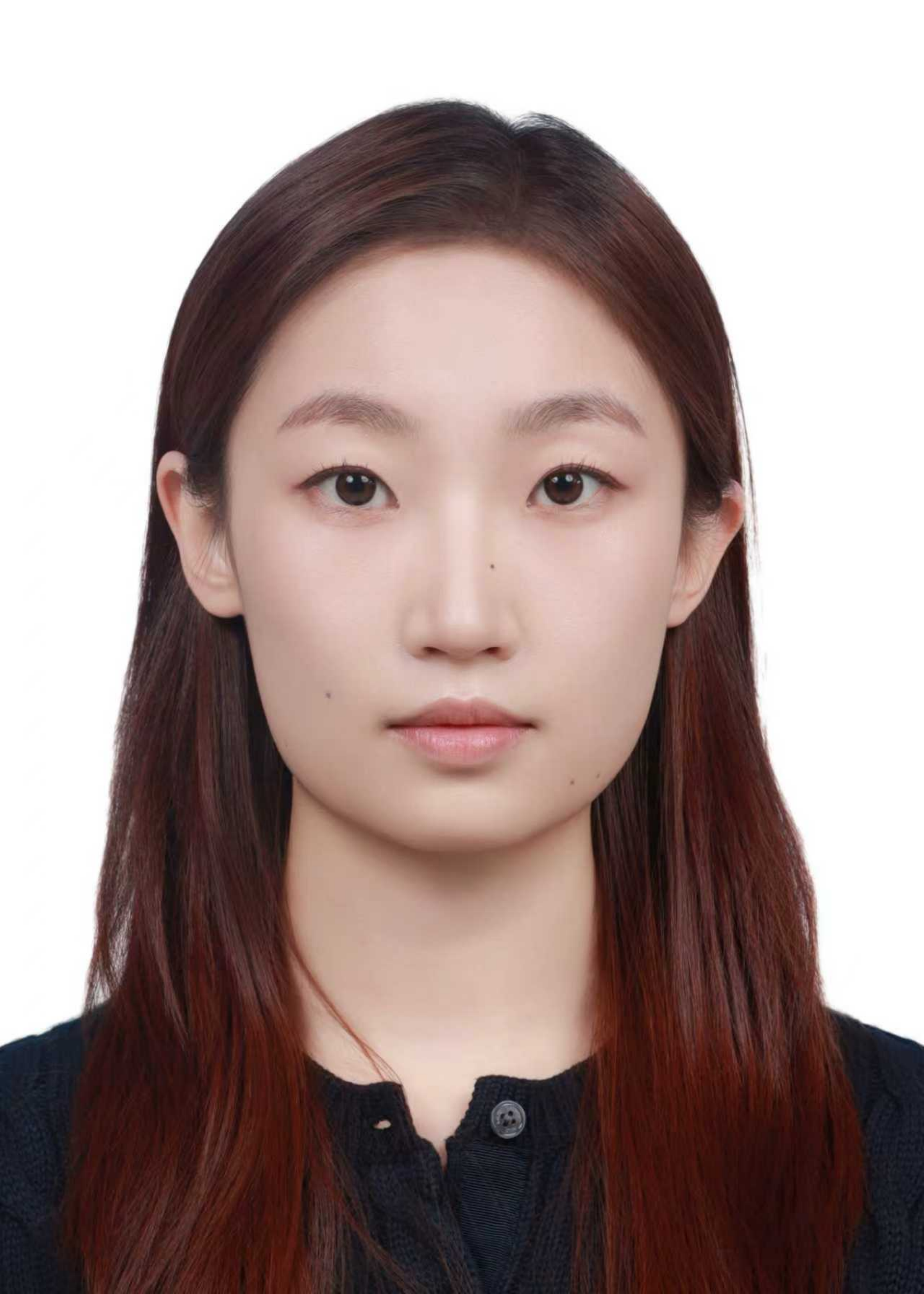}}]{Yue Fan} 
received the B.S. degree in Software Engineering from China University of Geosciences (Beijing), Beijing, China, in 2023. She is currently working toward the M.S. degree in mechanical engineering with the School of Advanced Manufacturing and Robotics, Peking University, Beijing, China. 

Her research interests include aerial vehicles, deep learning, and robot learning.
\end{IEEEbiography}

\begin{IEEEbiography}[{\includegraphics[width=1in,height=1.25in,clip,keepaspectratio]{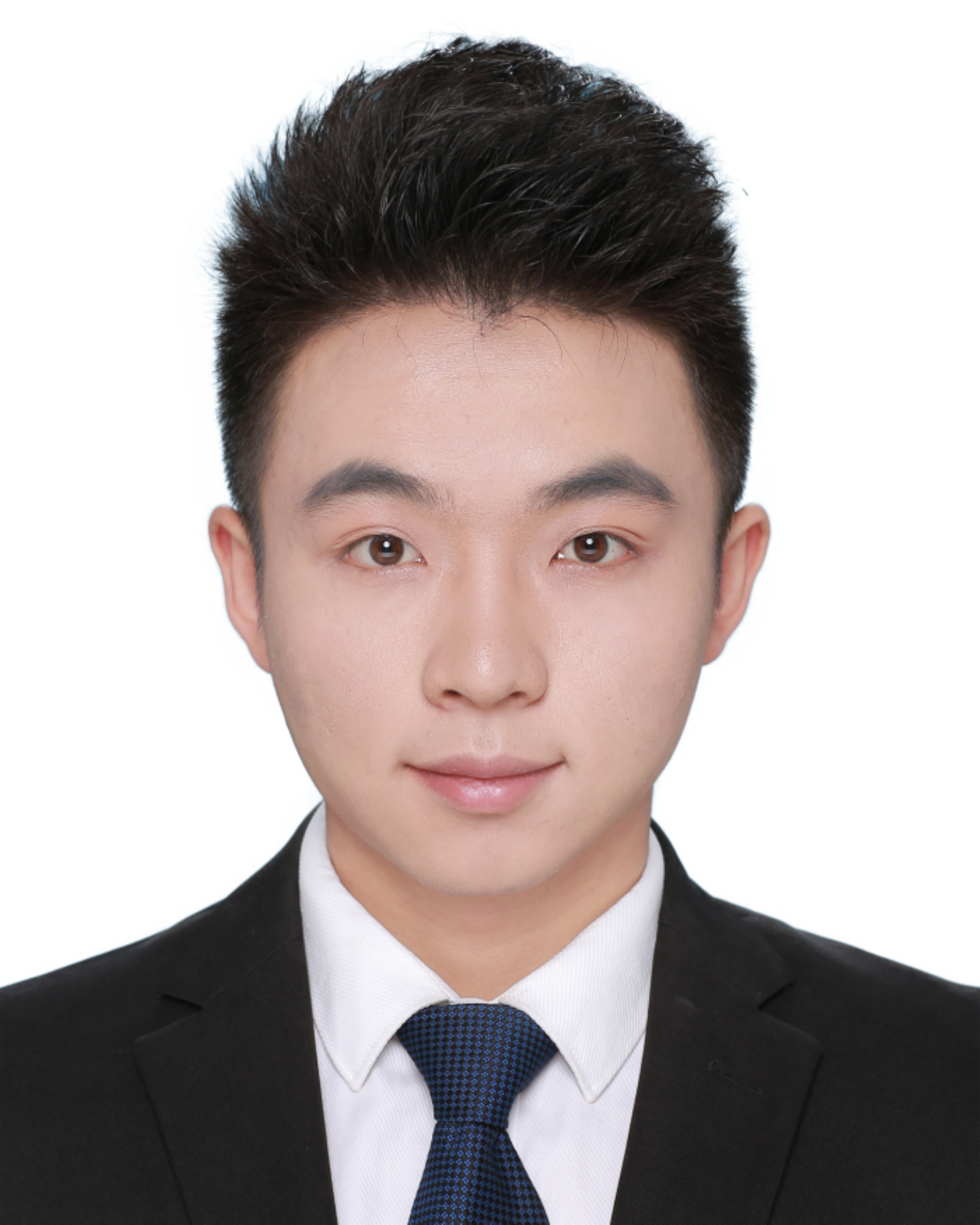}}]{Hao Cheng} (Student Member, IEEE) 
received the Bachelor's degree in New Energy Science and Engineering (Wind Power) from North China Electric Power University, China, in 2017, and the Master's degree in Control Engineering from Tsinghua University, China, in 2021. 
He is currently working toward the Ph.D. degree in general mechanics and foundation of mechanics with the School of Advanced Manufacturing and Robotics, Peking University, Beijing, China. 

His research interests include \mbox{lighter-than-air} aerial vehicles, bio-inspired robotics, and continuum robots. 
\end{IEEEbiography}

\begin{IEEEbiography}[{\includegraphics[width=1in,height=1.25in,clip,keepaspectratio]{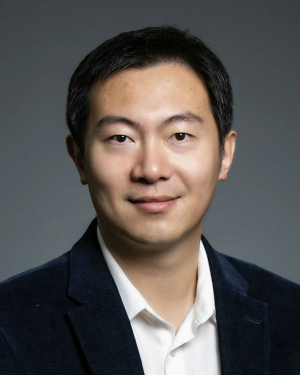}}]{Feitian Zhang} (Member, IEEE) 
received the bachelor's and master's degrees in automatic control from the Harbin Institute of Technology, Harbin, China, in 2007 and 2009, respectively, and the Ph.D. degree in electrical and computer engineering from Michigan State University, East Lansing, MI, USA, in 2014.

He was a Postdoctoral Research Associate with the Department of Aerospace Engineering and Institute for Systems Research, University of Maryland, College Park, MD, USA, from 2014 to 2016, and an Assistant Professor of Electrical and Computer Engineering with George Mason University, Fairfax, VA, USA, from 2016 to 2021. He is currently an Associate Professor of Robotics Engineering with Peking University, Beijing, China. His research interests include mechatronics systems, robotics and controls, aerial vehicles, and underwater vehicles. 
\end{IEEEbiography}

\end{document}